\documentclass[10pt,twocolumn,letterpaper]{article}

\usepackage[pagenumbers]{cvpr} %

\usepackage[dvipsnames]{xcolor}

\definecolor{tabfirst}{rgb}{1, 0.7, 0.7}
\definecolor{tabsecond}{rgb}{1, 0.85, 0.7}
\definecolor{tabthird}{rgb}{1, 1, 0.7}

\newcommand{\figref}[1]{Fig.~\ref{fig:#1}}
\newcommand{\tabref}[1]{Table~\ref{tab:#1}}
\newcommand{\eqnref}[1]{Eqn.~\ref{eqn:#1}}
\newcommand{\secref}[1]{Sec.~\ref{sec:#1}}

\def\eg{\emph{e.g}\onedot}

\def\ie{\emph{i.e}\onedot}

\newcommand{\inputImages}{x^{\mathrm{obs}}}
\newcommand{\inputCameras}{\pi^{\mathrm{obs}}}
\newcommand{\targetImage}{x}
\newcommand{\targetCamera}{\pi}
\newcommand{\pixelnerf}{R_{\phi}}
\newcommand{\encoder}{\mathcal{E}}
\newcommand{\unet}{\mathcal{\epsilon_\theta}}
\newcommand{\decoder}{\mathcal{D}}

\newcommand{\latentT}{z_t}
\newcommand{\nerfparams}{\psi}
\newcommand{\featuremap}{f}

\definecolor{tableyellow}{rgb}{1, 1, 0.7}
\definecolor{tableorange}{rgb}{1, 0.85, 0.7}
\definecolor{tablered}{rgb}{1, 0.7, 0.7}

\newcommand\lft{\mathopen{}\left}
\newcommand\rgt{\aftergroup\mathclose\aftergroup{\aftergroup}\right}

\newcommand{\norm}[1]{\lft\lVert#1\rgt\rVert}

\newcommand{\downsample}[1]{#1_{\downarrow}}

\newcommand{\myparagraph}[1]{ \vspace{3pt}  \noindent {\bf #1}\,\,\,}
\newcommand{\zipnerf}{$\text{Zip-NeRF}^*$}
\newcommand{\zeronvs}{$\text{ZeroNVS}^*$}

\newcommand{\ours}{ReconFusion\xspace}
\newcommand{\Ours}{ReconFusion\xspace} %

\definecolor{cvprblue}{rgb}{0.21,0.49,0.74}
\usepackage[pagebackref,breaklinks,colorlinks,citecolor=cvprblue]{hyperref}
\usepackage{multirow}
\usepackage{tabu}
\usepackage{xcolor,colortbl}

\def\paperID{8641} %
\def\confName{CVPR}
\def\confYear{2024}

\title{ReconFusion: 3D Reconstruction with Diffusion Priors}

\makeatletter
\newcommand{\printfnsymbol}[1]{%
        \textsuperscript{\@fnsymbol{#1}}%
}
\makeatother

\author{
Rundi Wu$^{1,2}$\footnote{}
\quad
Ben Mildenhall$^{1}$\printfnsymbol{1}
\quad
Philipp Henzler$^1$
\quad
Keunhong Park$^1$
\quad
Ruiqi Gao$^3$
\quad
Daniel Watson$^3$
\\
Pratul P. Srinivasan$^1$
\quad
Dor Verbin$^1$
\quad
Jonathan T. Barron$^1$
\quad
Ben Poole$^3$
\quad
Aleksander Hołyński$^{1}$\printfnsymbol{1}\vspace{0.75em}
\\
\centerline{$^1$Google Research \quad $^2$Columbia University\quad $^3$Google DeepMind}\\\vspace{0.75em}
\centerline{\small\printfnsymbol{1} denotes equal contribution}\vspace{-2.45em}
}

\usepackage{tabularray}

\begin{document}

\newcommand{\teaserwide}{0.9in}

\twocolumn[{%
\renewcommand\twocolumn[1][]{#1}%
\maketitle

\centering
\resizebox{\textwidth}{!}{

\begin{tblr}{
  colspec = {X[c,m,0.9] X[c,m,0.15] X[c] X[c] X[c]},
  rowspec = {Q[m,rowsep=0pt] Q[m,rowsep=0pt] Q[m,rowsep=2pt]},
  stretch = 0pt,
  colsep = 0pt,
  hlines = {white, 1pt},
  vlines = {white, 1pt},
}
    \SetCell[r=3]{c,m} \includegraphics[width=\linewidth]{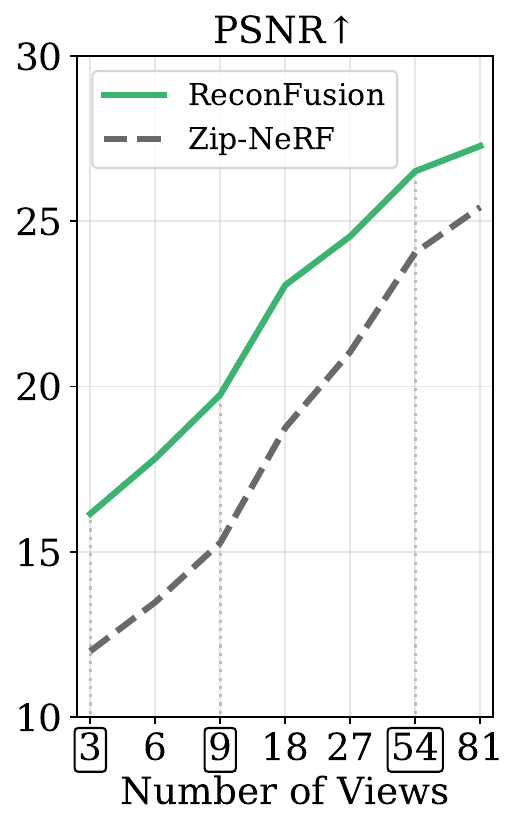} &
    \rotatebox{90}{\begin{minipage}{2.7cm}\centering \Ours \end{minipage}} &
    \includegraphics[width=1.0\linewidth]{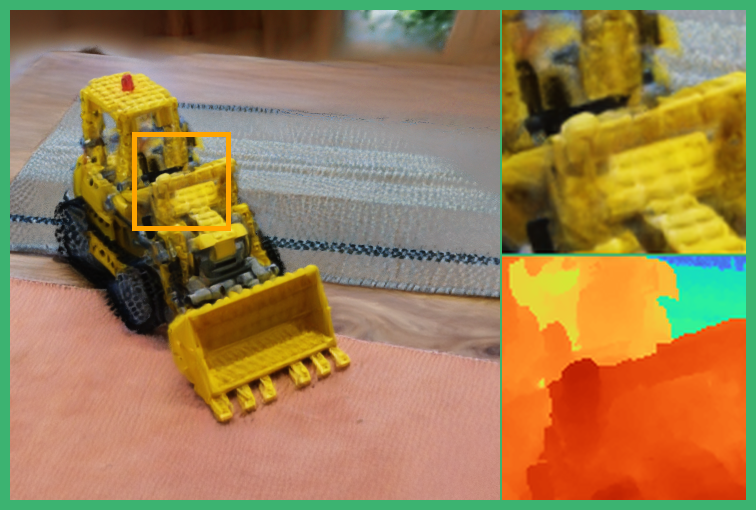} &
    \includegraphics[width=1.0\linewidth]{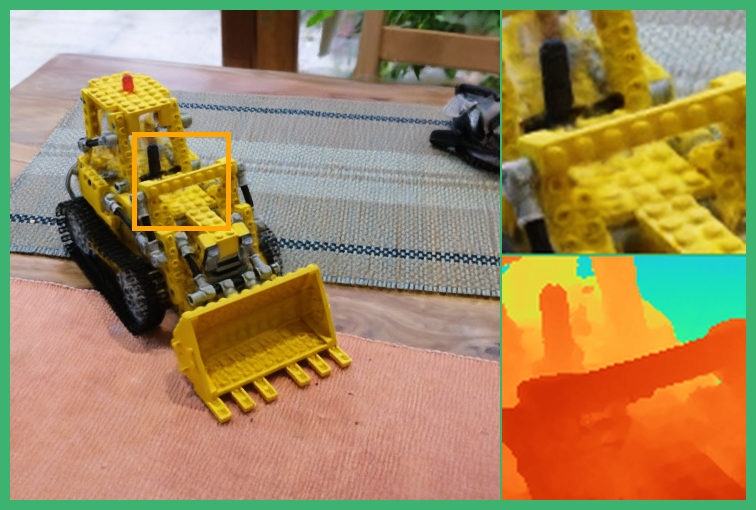} &
    \includegraphics[width=1.0\linewidth]{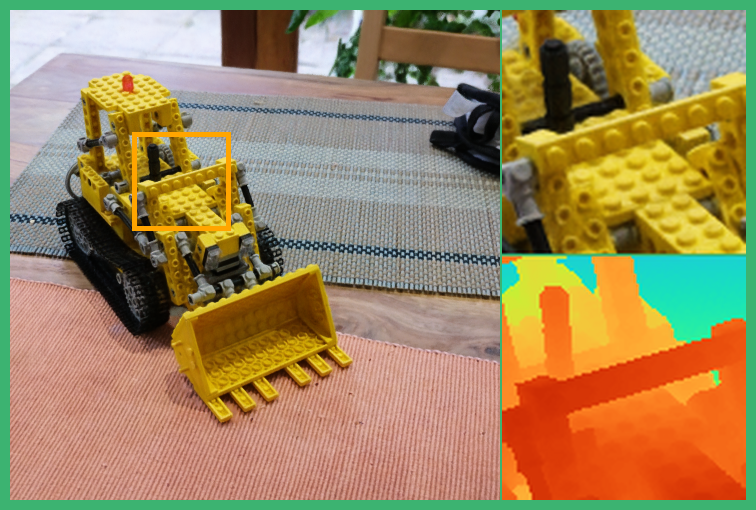} \\
    & \rotatebox{90}{\begin{minipage}{2.7cm}\centering Zip-NeRF~\cite{barron2023zip}\end{minipage}} &
    \includegraphics[width=1.0\linewidth]{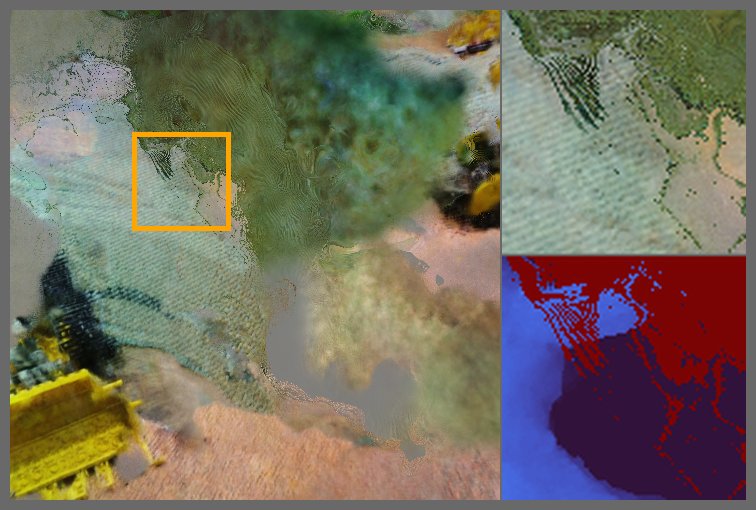} &
    \includegraphics[width=1.0\linewidth]{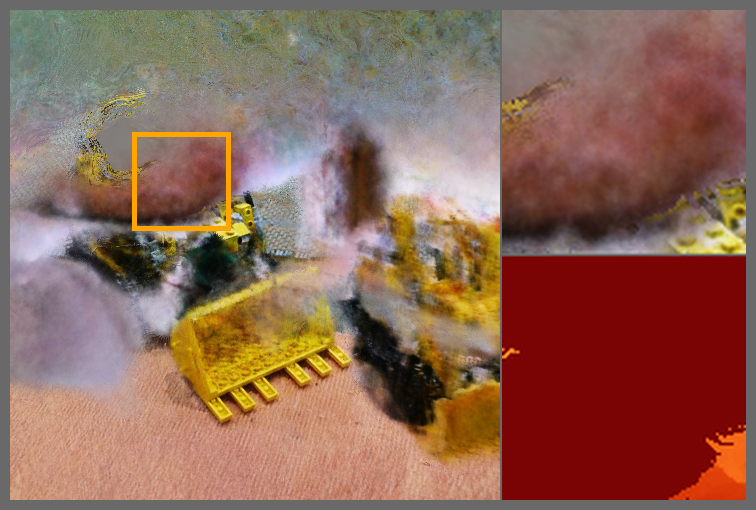} &
    \includegraphics[width=1.0\linewidth]{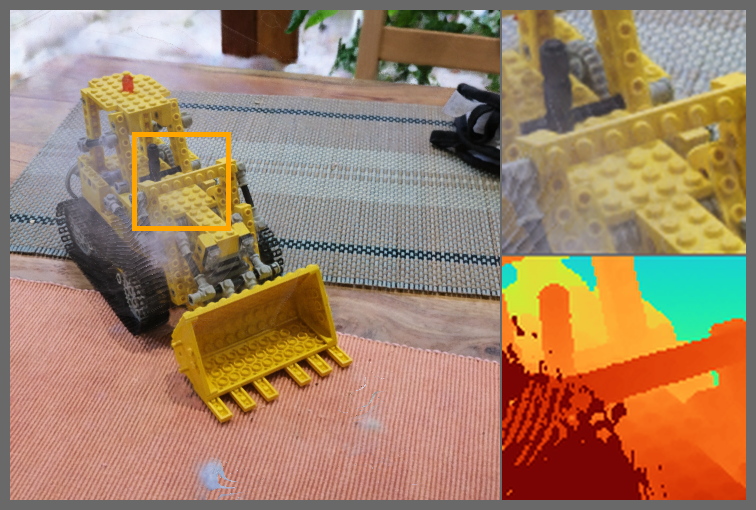} \\
    & & \small \small 3 views 
    & \small \small 9 views 
    & \small 54 views \\
\end{tblr}
}

\captionof{figure}{
Methods for reconstructing a 3D scene from images, such as Neural Radiance Fields (NeRF), often exhibit artifacts when trained with few input views.
\ours uses a diffusion model trained for novel view synthesis to regularize NeRF optimization. When the reconstruction problem is severely underconstrained (3 and 9 views), this prior can greatly improve robustness and often prevent catastrophic failures. Even in the case of significantly more observations (54 views), \ours improves quality and helps reduce ``floater'' artifacts common to volumetric reconstruction methods like NeRF.
We encourage the reader to view the results at \href{https://reconfusion.github.io}{reconfusion.github.io} to see the improvement our method can provide in few-view captures of real world scenes.
\phantom{euler bojangles phantom text to add some vspace below the caption, blah blah blah blah blah blah blah blah blah}
}
\label{fig:teaser}
}]

\maketitle
\begin{abstract}
\vspace{-3mm}
3D reconstruction methods such as Neural Radiance Fields (NeRFs) excel at rendering photorealistic novel views of complex scenes. However, recovering a high-quality NeRF typically requires tens to hundreds of input images, resulting in a time-consuming capture process. 
We present \ours to reconstruct real-world scenes using only a few photos. Our approach leverages a diffusion prior for novel view synthesis, trained on synthetic and multiview datasets, which regularizes a NeRF-based 3D reconstruction pipeline at novel camera poses beyond those captured by the set of input images.
Our method synthesizes realistic geometry and texture in underconstrained regions while preserving the appearance of observed regions. 
We perform an extensive evaluation across various real-world datasets, including forward-facing and 360-degree scenes, demonstrating significant performance improvements over previous few-view NeRF reconstruction approaches.
Please see our project page at \href{https://reconfusion.github.io/}{reconfusion.github.io}.
\end{abstract}
\vspace{-5mm}
\section{Introduction}
\label{sec:intro}
Advances in 3D reconstruction have enabled the transformation of images of real-world scenes into 3D models which produce photorealistic renderings from novel viewpoints~\cite{mildenhall2020nerf,kerbl3Dgaussians}.
Methods like NeRF~\cite{mildenhall2020nerf} optimize a 3D representation whose renderings match observed input images at given camera poses. However, renderings from under-observed views are artifact-prone, particularly in less densely captured areas. As such, recovering a high-quality NeRF requires exhaustive scene capture, where each region is photographed from multiple angles multiple times.

NeRF's dense capture requirement poses a major challenge, necessitating tens to hundreds of images for even simple objects to ensure a clean reconstruction (\figref{teaser}).
Many methods aim to reduce the reliance on dense captures
by developing heuristic low-level regularizers based on reconstructed depth~\cite{deng2022depth,roessle2022dense,wang2023sparsenerf}, visibility~\cite{somraj2023vip,kwak2023geconerf}, appearance~\cite{niemeyer2022regnerf,wynn2023diffusionerf}, or image-space frequencies~\cite{yang2023freenerf}. However, even the most effective methods show considerable degradation at novel viewpoints compared to denser captures. 

With the recent success of diffusion models for image generation~\cite{ho2020denoising,song2020denoising}, researchers have applied diffusion models to the task of novel view synthesis --- modeling the distribution of unseen views given observations from known views~\cite{watson2022novel,gu2023nerfdiff,liu2023zero}. While these models excel at generating realistic images from novel view points, they do not produce a single consistent 3D shape from a sparse set of input views. Existing work produces 3D models that are either trained per category~\cite{zhou2023sparsefusion,chan2023generative,tewari2023diffusion,yoo2023dreamsparse, gu2023nerfdiff}, or are limited to single image inputs containing an object~\cite{liu2023zero,liu2023syncdreamer,shi2023mvdream}, preventing their use as a general prior for 3D scene reconstruction.

Our proposed method uses 2D image priors over novel views to enhance 3D reconstruction. We derive this prior from a diffusion model trained for novel view synthesis.
Given multiple posed images of a scene, this model estimates the scene's appearance from novel viewpoints.
As posed multiview data is limited (compared to massive single image datasets), we finetune our diffusion model from a pretrained latent diffusion model~\cite{rombach2022high} on a mixture of real world and synthetic multiview image datasets: RealEstate10K~\cite{zhou2018stereo}, CO3D~\cite{reizenstein2021common}, MVImgNet~\cite{yu2023mvimgnet}, and Objaverse~\cite{deitke2023objaverse}. Once trained, this model is used to regularize a typical NeRF reconstruction pipeline by using an approach similar to score distillation sampling (SDS)~\cite{poole2022dreamfusion}.

Our approach outperforms existing baselines on several datasets of both forward-facing and unbounded 360$^{\circ}$ scenes.
Furthermore, we show that our diffusion prior is an effective drop-in regularizer for NeRFs across a range of capture settings.
In few-view scenarios with limited scene observations, it provides a strong prior for plausible geometry and appearance reconstruction.
In denser capture settings, it helps reduce distracting ``fog'' and ``floater'' artifacts while preserving the appearance of well-sampled regions.

Many aspects of our pipeline have been explored in prior work. We contribute an end-to-end system that markedly improves 3D reconstruction quality, uniquely combining the challenges of developing a multiview-conditioned image diffusion model and integrating it into the NeRF optimization process, minimizing the need for rigorous capture.

\section{Related Work}
\label{sec:relatedwork}

\myparagraph{Few-view NeRF}
Minimizing NeRF's need for dense capture is crucial for democratizing 3D capture, and has motivated many works ~\cite{jain2021putting,deng2022depth,roessle2022dense,somraj2023vip,somraj2023simplenerf,niemeyer2022regnerf,wynn2023diffusionerf,yang2023freenerf,wang2023sparsenerf,kwak2023geconerf,seo2023mixnerf,warburg2023nerfbusters}.
Most existing methods focus on regularizing the geometry of the scene. DS-NeRF~\cite{deng2022depth} 
utilizes sparse depth outputs from Structure-from-Motion (SfM) as supervision. DDP-NeRF~\cite{roessle2022dense} further uses a CNN to obtain dense depth supervision from sparse inputs. SimpleNeRF~\cite{somraj2023simplenerf} regularizes appearance and geometry by training two additional models which respectively reduce positional encoding frequencies and remove view-dependent components.
Similarly, FreeNeRF~\cite{yang2023freenerf} demonstrates that simply regularizing the frequency range of NeRF's positional encoding features improves quality in few-view scenarios.
RegNeRF~\cite{niemeyer2022regnerf} introduces a depth smoothness loss and a pre-trained normalizing flow color model to regularize the geometry and appearance of novel views.
DiffusioNeRF~\cite{wynn2023diffusionerf} trains a diffusion model to regularize the distribution of RGB-D patches from perturbed viewpoints.
GANeRF~\cite{roessle2023ganerf} trains a generator network to improve NeRF renderings and an image discriminator network to provide feedback that can be used to improve the reconstruction in a multiview-consistent manner.
While all these methods regularize ambiguous geometry and appearance during NeRF optimization, they often fail on larger scenes when the view sparsity is extreme.

\myparagraph{Regression models for view synthesis}
While NeRFs are optimized \emph{per-scene}, other methods train feed-forward neural networks for generalized novel view synthesis. These networks leverage large collections of posed multiview data across many scenes~\cite{yu2021pixelnerf,flynn2016deepstereo,zhou2018stereo,chen2021mvsnerf,wang2021ibrnet,tucker2020single,trevithick2021grf,sajjadi2022scene,henzler2021wct}.
Most methods lift the input images into a 3D representation, like using a plane sweep volume, and predict novel views in a feed-forward manner.
They work well near the input views, but extrapolate poorly to ambiguous views, where the distribution of possible renderings becomes multi-modal.

\begin{figure*}
    \centering
    \includegraphics[trim={0 28.3cm 38.8cm 0},clip,width=\linewidth]{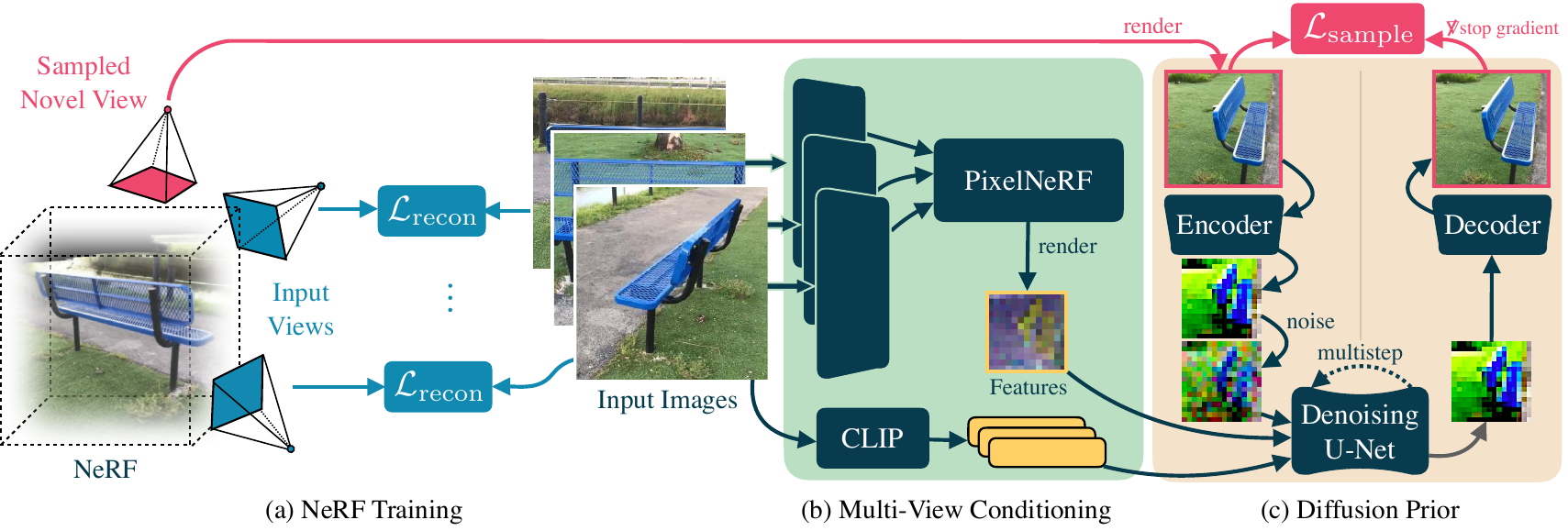}
    \caption{
    (a) We optimize a NeRF to minimize a reconstruction loss $\mathcal{L}_{\mathrm{recon}}$ between renderings and a limited set of input images, alongside a sample loss $\mathcal{L}_{\mathrm{sample}}$ comparing renderings from random poses and with predictions by a diffusion model for those poses.
    (b)~To generate the sample image, we use a PixelNeRF-style model~\cite{yu2021pixelnerf} to fuse input image data, rendering a feature map for the sample view.
    (c) This feature map, merged with the noisy latent (computed by adding some amount of noise to the current NeRF rendering from that pose), is provided to a diffusion model. This model additionally uses CLIP embeddings of the input images via cross-attention, generating a decoded output sample. This sample is used to apply an image-space loss to the corresponding NeRF rendering.
    }
    \label{fig:overview}
\end{figure*}

\myparagraph{Generative models for view synthesis}
Extrapolating beyond observed inputs for view synthesis requires generating unknown parts of the scene.
Earlier works addressing this problem primarily leverage Generative Adversarial Networks (GANs)~\cite{goodfellow2014generative,gadelha20173d,zhu2018visual,nguyen2019hologan,henzler2019platonicgan,niemeyer2021giraffe,chan2021piGAN,chan2022eg3d,gao2022get3d,schwarz2022voxgraf}. More recent works use diffusion models, following their immense success on image generation~\cite{wang2023sparsenerf,chan2023generative,gu2023nerfdiff,ren2022look,tewari2023diffusion,yoo2023dreamsparse,liu2023zero,tseng2023consistent,liu2023syncdreamer,shi2023mvdream,karnewar2023holofusion,karnewar2023holodiffusion}.
3DiM~\cite{watson2022novel} trains a pose-conditioned image-to-image diffusion model on synthetic ShapeNet data~\cite{chang2015shapenet}.
GeNVS~\cite{chan2023generative} and SparseFusion~\cite{wang2023sparsenerf} train on real-world multiview data~\cite{reizenstein2021common} and further incorporate 3D geometry priors by conditioning on rendered features~\cite{yu2021pixelnerf,suhail2022generalizable}.
While they show promising results for novel view synthesis, their models are category-specific and do not generalize to arbitrary scenes.
Recently, Zero-1-to-3~\cite{liu2023zero} fine-tunes a large-scale pretrained diffusion model~\cite{rombach2022high} on the synthetic Objaverse dataset~\cite{deitke2023objaverse} and achieves strong zero-shot generalization on real images.
However, it only supports images of objects with clean backgrounds (versus full real scenes) and is limited to single-image inputs.
ZeroNVS~\cite{sargent2023zeronvs} further fine-tunes Zero-1-to-3 to enable single-image reconstruction of general scenes.
Our approach is similar, but utilizes a PixelNeRF-based approach~\cite{yu2021pixelnerf} for conditioning (similar to GeNVS~\cite{chan2023generative}) to allow for any number of input images and provide more precise pose conditioning, and fine-tunes a pretrained image diffusion model on real-world multiview datasets~\cite{reizenstein2021common,zhou2018stereo,yu2023mvimgnet}, which combined facilitates few-view reconstruction on arbitrary scenes.

\myparagraph{Lifting 2D diffusion models for 3D generation}
Given the limited amount of 3D data available for training, recent works have attempted to leverage 2D diffusion models to generate 3D assets from a text prompt~\cite{poole2022dreamfusion,lin2023magic3d,wang2023score} or an input image~\cite{liu2023zero,shi2023mvdream}.
DreamFusion~\cite{poole2022dreamfusion} proposed score distillation sampling (SDS), where a 2D diffusion model acts as a critic to supervise the optimization of a 3D model. SparseFusion~\cite{zhou2023sparsefusion} proposes multistep sampling where an image is sampled given a noisy encoding of the current rendering as a target for 3D reconstruction.  We experiment with both approaches in our reconstruction pipeline.

\section{\Ours}
\Ours consists of a diffusion model trained for novel view synthesis and a 3D reconstruction procedure to make use of this diffusion model. We describe the details of the diffusion model training in \secref{diffusion} and how we use the diffusion model as a prior for 3D reconstruction in \secref{distill}. 

\subsection{Diffusion Model for Novel View Synthesis}
\label{sec:diffusion}
Given a set of posed images, we seek to learn a prior that can generate plausible novel views. If we can learn what the back or side of an object looks like given images of the front, we can use this to guide a 3D reconstruction process to recover a plausible 3D scene. Formally, we are given a set of input images $\inputImages\!=\!\{x_i\}_{i=1}^N$, corresponding camera parameters $\inputCameras\!=\!\{\pi_i\}_{i=1}^N$, and a target camera for a novel view $\pi$, and want to learn the conditional distribution over the image $x$ at the novel view:
$p\lft(\targetImage | \inputImages, \inputCameras, \targetCamera \rgt)$.

\myparagraph{Diffusion Models} We build on latent diffusion models (LDMs) \cite{rombach2022high} for their ability to efficiently model high resolution images. LDMs encode input images to a latent representation using a pretrained variational auto-encoder (VAE) $\encoder$. Diffusion is performed on these latents, where a denoising U-Net $\unet$ maps noisy latents back to clean latents. During inference, this U-Net is used to iteratively denoise pure Gaussian noise to a clean latent. To recover an image, the latents are passed through a VAE decoder $\decoder$.

\myparagraph{Conditioning} 
Similar to Zero-1-to-3~\cite{liu2023zero}, we start from an LDM trained for text-to-image generation, and additionally condition on input images and poses.
Converting a text-to-image model into a posed images-to-image model requires augmenting the U-Net architecture with additional conditioning pathways. 
To modify the pretrained architecture for novel view synthesis from multiple posed images, we inject two new conditioning signals into the U-Net (see \figref{overview}(b)). For high-level semantic information about the inputs, we use the CLIP~\cite{radford2021learning} embedding of each input image (denoted $e^\text{obs}$) and feed this sequence of feature vectors into the U-Net via cross-attention. For relative camera pose and geometric information, we use a PixelNeRF~\cite{yu2021pixelnerf} model $\pixelnerf$ to render a feature map $\featuremap$ with the same spatial resolution as the latents from the target viewpoint $\targetCamera$:
\begin{equation}
    \featuremap = \pixelnerf\lft(\inputImages, \inputCameras, \targetCamera \rgt)\,.
\end{equation}
This rendered feature map $\featuremap$ is a spatially aligned conditioning signal which implicitly encodes the relative camera transform.
We concatenate $\featuremap$ with the noisy latent along the channel dimension, and feed it into the denoising U-Net $\unet$. This feature map conditioning strategy is similar to the one used in GeNVS~\cite{chan2023generative}, SparseFusion~\cite{zhou2023sparsefusion}, and other recent works to better provide an accurate representation of the novel camera pose, as compared to attending over a direct embedding of the camera extrinsics and intrinsics themselves (an ablation study can be found in \secref{ablation}).

\myparagraph{Training}
 We freeze the weights of the pretrained encoder and decoder, initialize the U-Net parameters $\theta$ from pretrained weights, and optimize the modified architecture for view synthesis using the simplified diffusion loss \cite{ho2020denoising}:
\begin{small}
\begin{align}
    \mathcal{L}_\mathrm{Diff}(\theta, \phi) &= \mathbb{E}_{\targetImage, \pi, \inputImages,\inputCameras, \epsilon,t} \norm{ \epsilon - \unet(\latentT, t, e^\text{obs}, f)}^2 \,,
\label{eqn:loss_diffusion}
\end{align}
\end{small}
where $t\in\{1, \ldots, T\}$ is the diffusion timestep, $\epsilon\sim\mathcal{N}(0,I)$,  $\latentT=\alpha_t\encoder(\targetImage)+\sigma_t\epsilon$ is the noisy latent at that timestep,  $e^\text{obs}$ are the CLIP image embeddings for the input images $\inputImages$, and $f$ is the rendered feature map from PixelNeRF $R_\phi$. 
In addition to the loss in \eqnref{loss_diffusion}, we optimize the PixelNeRF parameters $\phi$ with a photometric loss:
\begin{equation}
    \mathcal{L}_\mathrm{PixelNeRF}(\phi) = \mathbb{E}_{\inputImages, \inputCameras, \targetImage, \targetCamera} \norm{c - \downsample{\targetImage}}^2\,,
\end{equation}
where $c$ is an output of the PixelNeRF model (at the same resolution as the feature map $f$) and $\downsample{\targetImage}$ is the target image downsampled to the spatial resolution of $\latentT$ and $f$.
This loss encourages PixelNeRF to reconstruct the RGB target image, which helps to avoid bad local minima where the diffusion model is unable to leverage the PixelNeRF inputs.

Due to the use of cross-attention and the design of PixelNeRF, both conditioning branches can take an arbitrary number and permutation of input images. This enables the model to be trained and evaluated with a variable number of observed posed images.
While there are many ways to condition on images and poses, we found our design more effective than alternatives (see the ablation in \secref{ablation}).

\subsection{3D Reconstruction with Diffusion Priors}
\label{sec:distill}
The trained diffusion model produces plausible single images for novel camera poses, but generated images are often inconsistent for different poses or random seeds.
State-of-the-art NeRF methods produce 3D consistent 3D models, but often exhibit volumetric ``floater'' artifacts and inaccurate (or totally unrecognizable) geometry from novel views.
To enable 3D reconstruction from a smaller number of posed inputs, we augment the state-of-the-art 3D reconstruction pipeline from Zip-NeRF~\cite{barron2023zip} with a prior from our diffusion model trained for novel view synthesis.

\myparagraph{Reconstruction loss} NeRF-based methods optimize a randomly initialized 3D model to match a set of posed images. 
The NeRF parameters $\nerfparams$ are optimized by minimizing the reconstruction error between a rendered image $x=x(\nerfparams,\inputCameras)$ and an observed image $x^\text{obs}$ at pose $\inputCameras$:
\begin{equation}
\mathcal{L}_\text{Recon}(\nerfparams) = \mathbb{E}_{\inputImages,\inputCameras}\left[\ell(x(\nerfparams, \inputCameras), \inputImages)\right],
\end{equation}
where $\ell$ is an image similarity loss function such as the $\ell_2$-norm or a robust loss. This loss is only evaluated where we have observations, and thus the training procedure never views the 3D model from novel views. 

\myparagraph{Diffusion loss}
In addition, we seek to optimize the 3D model to produce realistic rendering at novel views unobserved in the inputs.
To do so, we use a diffusion model which provides a prior on the distribution of plausible images of the scene.
We distill this prior into a consistent 3D model by using a regularization loss derived from the diffusion model outputs.
We experimented with several losses and discuss our multistep sampling approach below. 

At each iteration, we sample a random view and generate an image from the diffusion model to produce a target image (see \figref{overview}(a)). We can control how grounded the target image is to the current rendered image by starting the sampling process from an intermediate noise level. 
Specifically, we render an image $x(\nerfparams, \pi)$ from a sampled novel viewpoint $\pi$, and encode and perturb it to a noisy latent $\latentT$ with noise level $t \sim \mathcal{U}[t_\mathrm{min}, t_\mathrm{max}]$. 
We then generate a sample from the latent diffusion model by running DDIM sampling~\cite{song2020denoising} for $k$ intermediate steps, uniformly spaced between the smallest noise level and $t$, yielding a latent sample $z_0$. This latent is decoded to produce a target image $\hat x_{\pi} = \decoder\lft(z_0\rgt)$, which we use to supervise the rendering:
\begin{small}
\begin{equation}
\mathcal{L}_\mathrm{sample}(\nerfparams)= \mathbb{E}_{\pi, t} \big[w(t)\left( \norm{x - \hat x_{\pi}}_1 + \mathcal{L}_\mathrm{p}\lft(x, \hat x_{\pi}\rgt) \right)\big]\,,
\end{equation}
\end{small}
where $\mathcal{L}_\mathrm{p}$ is the perceptual distance LPIPS~\cite{zhang2018unreasonable}, and $w(t)$ is a noise-level dependent weighting function. This diffusion loss is most similar to SparseFusion~\cite{zhou2023sparsefusion}, and resembles the iterative dataset update strategy of Instruct-NeRF2NeRF~\cite{haque2023instruct}, except we sample a new image at each iteration. We empirically found this approach to work better than score distillation sampling~\cite{poole2022dreamfusion} (see supp. and \figref{ablation_3d}).

\paragraph{Novel view selection}
Which views should we sample when using our diffusion prior?
We do not want to place novel views inside of objects or behind walls, and the placement of views often depends on the scene content and type of capture. 
As in prior work such as RegNeRF~\cite{niemeyer2022regnerf}, we wish to define a distribution based on the known input poses and capture pattern that will encompass a reasonable set of novel camera poses, roughly matching the positions from which we would expect to observe the reconstructed scene.

We achieve this by determining a base set or path of poses through the scene, which we can randomly sample and perturb to define a full pose distribution for novel views.
In forward-facing captures such as LLFF~\cite{mildenhall2019local} and DTU~\cite{jensen2014large} or 360-degree captures such as mip-NeRF 360~\cite{barron2022mip}, we define an elliptical path fit to the training views, facing toward the focus point (the point with minimum average distance to the training cameras' focal axes).
In more unstructured captures such as CO3D~\cite{reizenstein2021common} and RealEstate10K~\cite{zhou2018stereo}, we fit a B-spline to roughly follow the trajectory of the training views. In either case, for each random novel view, we uniformly select one of the poses in the path and then perturb its position, up vector, and look-at point within some range. Please see the supplement for additional details.

\subsection{Implementation Details}
\label{sec:details}
Our base diffusion model is a re-implementation of the Latent Diffusion Model~\cite{rombach2022high} that has been trained on an internal dataset of image-text pairs with input resolution $512 \times 512 \times 3$ and a latent space with dimensions $64 \times 64\times 8$.
The encoder of our PixelNeRF is a small U-Net that takes as input an image of resolution $512 \times 512$ and outputs a feature map of resolution $64\times 64$ with $128$ channels (see the supplement for more details). 
We jointly train the PixelNeRF and finetune the denoising U-Net with batch size $256$ and learning rate $10^{-4}$ for a total of $250k$ iterations.
To enable classifier-free guidance (CFG), we set the input images to all zeros randomly with probability $10\%$.

We use Zip-NeRF~\cite{barron2023zip} as our backbone and train the NeRF for a total of $1000$ iterations.
The reconstruction term $\mathcal{L}_\mathrm{recon}$ uses the Charbonnier loss~\cite{charbonnier1994two} as in Zip-NeRF.
The weighting for $\mathcal{L}_\mathrm{sample}$ is linearly decayed from $1$ to $0.1$ over training, and the classifier-free guidance scale used for sampling is set to $3.0$.
We fix $t_\mathrm{max}=1.0$ for all training steps, and linearly anneal $t_\mathrm{min}$ from $1.0$ to $0.0$.
Regardless of $t$, we always sample the denoised image with $k=10$ steps.
In practice, diffusion models for view synthesis can be conditioned on a small number of observed input images and poses. Given a target novel view, we select the $3$ nearest camera positions from the observed inputs to condition the model. This enables our models to scale to large numbers of input images while selecting inputs that are most useful for the sampled novel view.
Please refer to the supplementary materials for more implementation details.

\section{Experiments}
\label{sec:experiments}
We evaluate \ours on five real-world datasets to demonstrate the performance and generalizability of our approach for few-view 3D reconstruction (\secref{results}).  We also perform several ablations on the components of the diffusion model and the 3D reconstruction procedure (\secref{ablation}). Finally, we demonstrate that our method improves NeRF reconstruction across a range of capture settings (\secref{scaling}). Additionally, we strongly advise the reader to view our supplementary video, as the visual difference in view synthesis results is significantly clearer in video than in still images.

\def \testdtu {figures/all_results/dtu_scan31_3_0014}
\def \testdtuB {figures/all_results/dtu_scan40_3_0001}
\def \testdtuC {figures/all_results/dtu_scan45_3_0002}

\def \testllffA {figures/all_results/llff_horns_3_0056}
\def \testllff {figures/all_results/llff_horns_3_0008}
\def \testllffC {figures/all_results/llff_fern_3_0016}
\def \testllffD {figures/all_results/llff_flower_3_0024}
\def \testllffE {figures/all_results/llff_leaves_3_0016}

\def \testcotd {figures/all_results/co3d_plant_188_20319_36755_6_0046}

\def \testretkA {figures/all_results/re10k_00e8df74b6805da7_3_0041}
\def \testretk {figures/all_results/re10k_00beb03ef95dc637_3_0022}
\def \testretkC {figures/all_results/re10k_000c3ab189999a83_3_0087}
\def \testretkD {figures/all_results/re10k_00e8df74b6805da7_3_0077}
\def \testretkE {figures/all_results/re10k_01a5cc3805e94c21_3_0022}
\def \testretkF {figures/all_results/re10k_01a5cc3805e94c21_3_0031}

\def \testmip {figures/all_results/mipnerf360_bicycle_9_0056}

\newcommand{\sixwide}{1.07in}
\newcommand{\sevenwide}{0.9in}
\begin{figure*}[t]
    \centering
        \begin{tabular}{@{}c@{\,\,}c@{\,\,}c@{\,\,}c@{\,\,}c@{\,\,}c@{\,\,}c@{\,\,}c@{}}
         & Zip-NeRF & DiffusioNeRF & FreeNeRF & SimpleNeRF & ZeroNVS & Ours & Ground Truth  \\
         \multirow{2}{*}[+2.5em]{\rotatebox{90}{ RealEstate10K (3)}}
         &
         \includegraphics[width=\sevenwide]{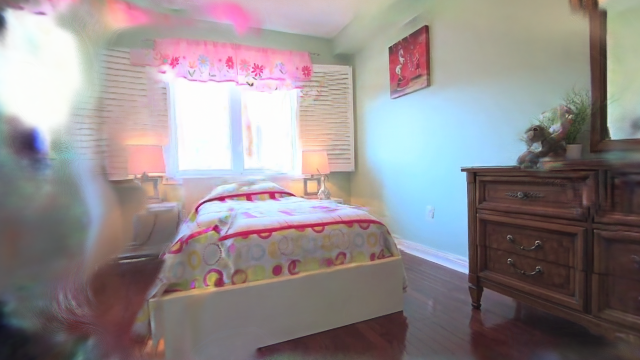} &
         \includegraphics[width=\sevenwide]{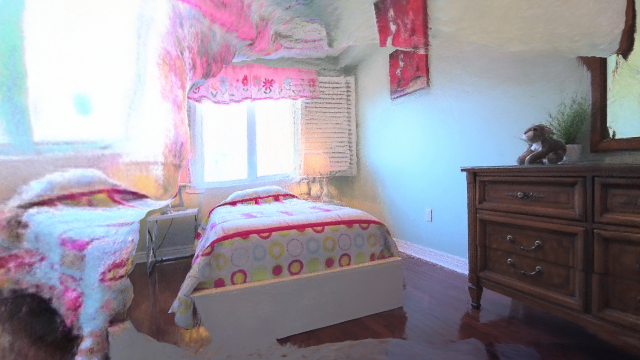} &
         \includegraphics[width=\sevenwide]{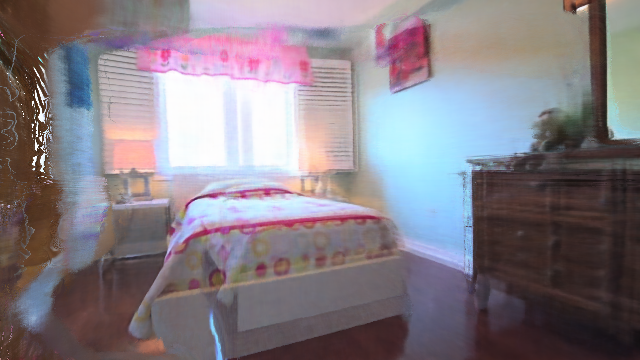} &
         \includegraphics[width=\sevenwide]{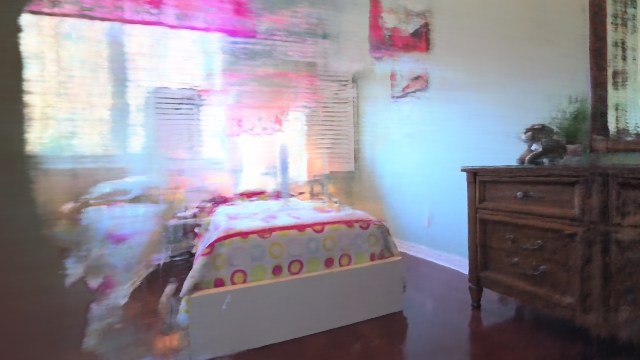} &
         \includegraphics[width=\sevenwide]{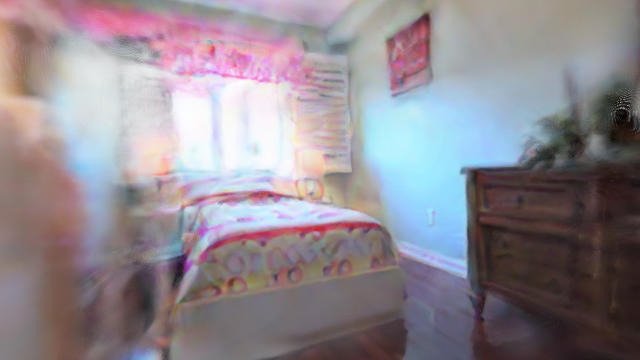} &
         \includegraphics[width=\sevenwide]{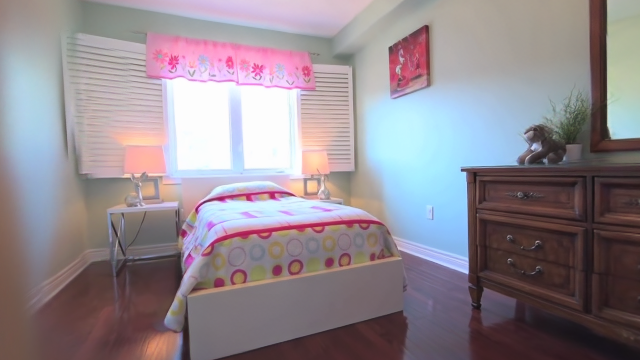} &
         \includegraphics[width=\sevenwide]{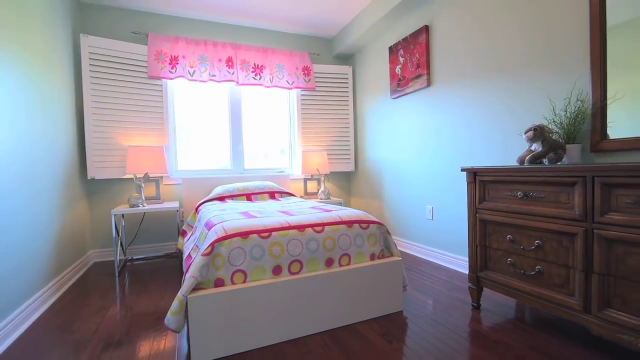} 
         \\
          & %
         \includegraphics[width=\sevenwide]{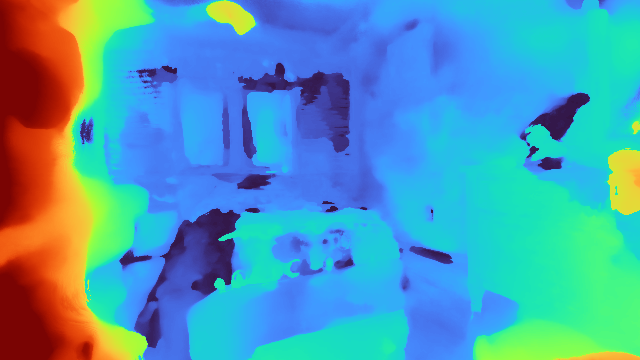} &
         \includegraphics[width=\sevenwide]{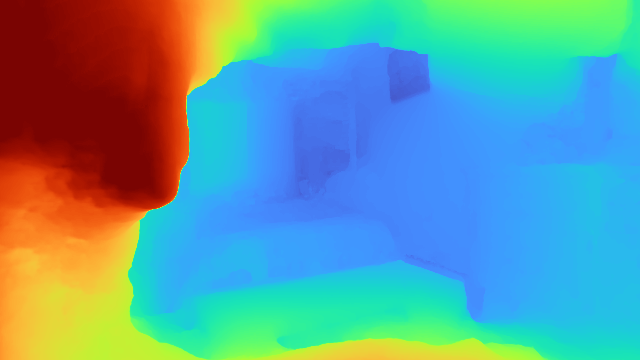} &
         \includegraphics[width=\sevenwide]{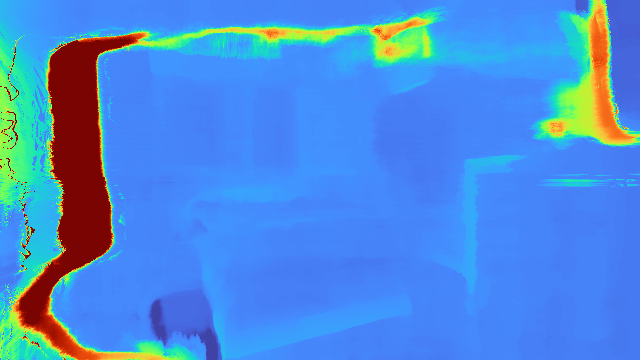} &
         \includegraphics[width=\sevenwide]{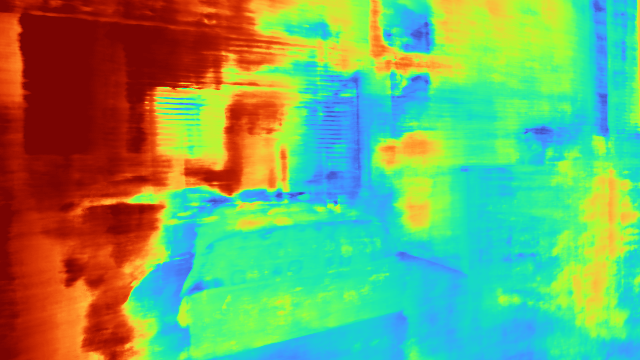} &
         \includegraphics[width=\sevenwide]{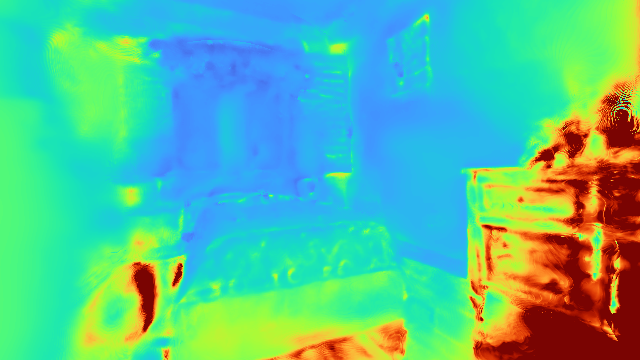} &
         \includegraphics[width=\sevenwide]{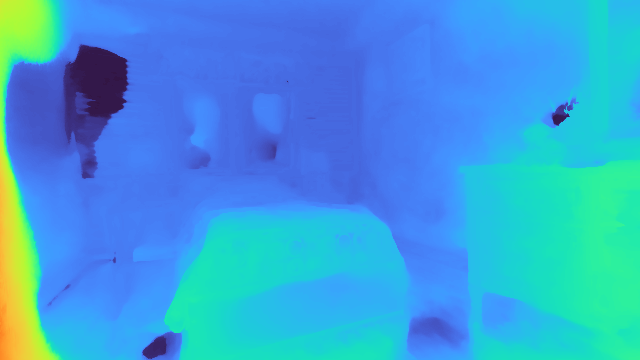} &
         \\
         \multirow{2}{*}[+1em]{\rotatebox{90}{ LLFF (3)}}
         &
         \includegraphics[width=\sevenwide]{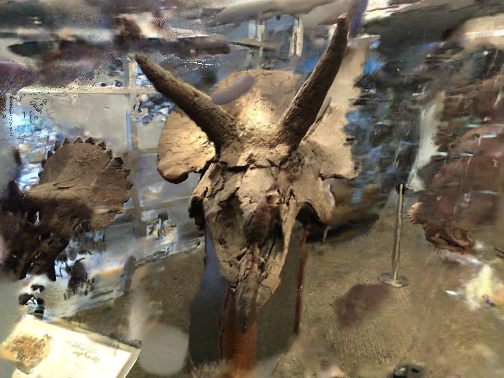} &
         \includegraphics[width=\sevenwide]{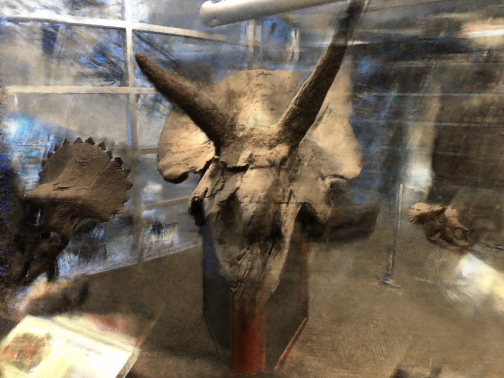} &
         \includegraphics[width=\sevenwide]{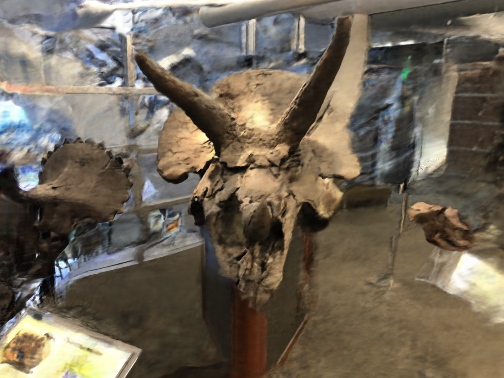} &
         \includegraphics[width=\sevenwide]{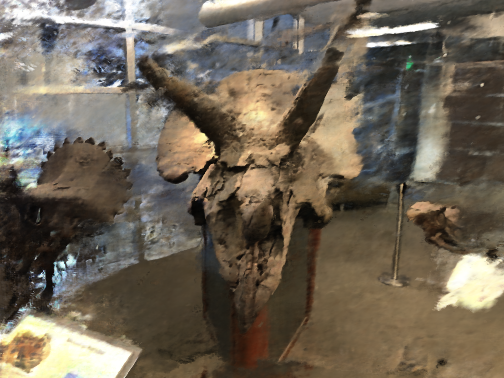} &
         \includegraphics[width=\sevenwide]{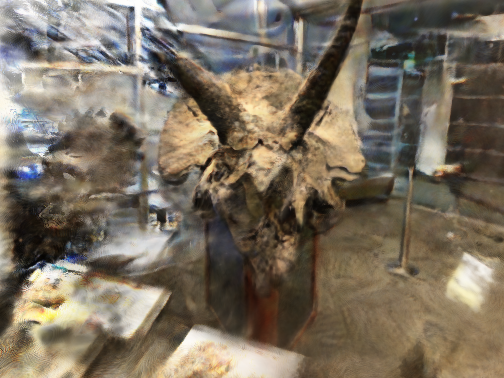} &
         \includegraphics[width=\sevenwide]{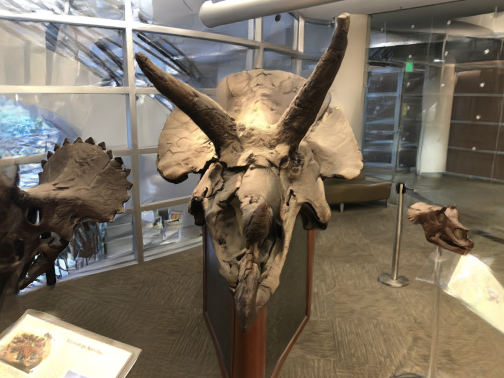} &
         \includegraphics[width=\sevenwide]{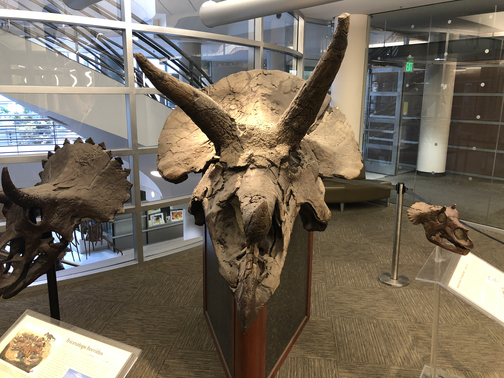} \\
          & %
         \includegraphics[width=\sevenwide]{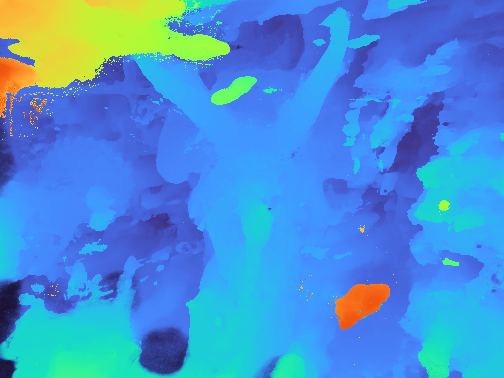} &
         \includegraphics[width=\sevenwide]{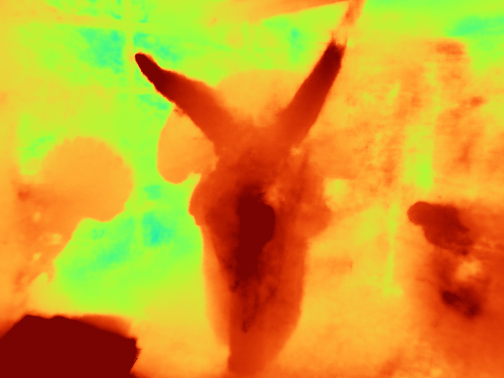} &
         \includegraphics[width=\sevenwide]{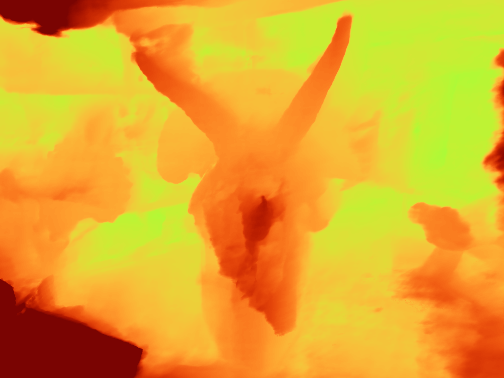} &
         \includegraphics[width=\sevenwide]{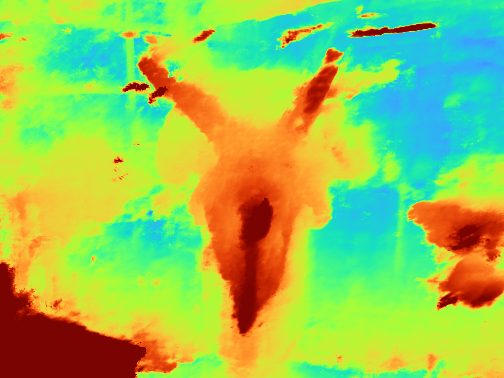} &
         \includegraphics[width=\sevenwide]{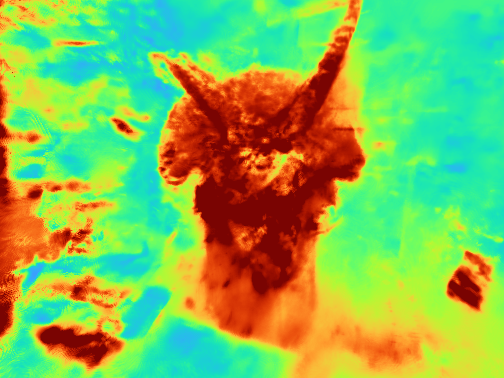} &
         \includegraphics[width=\sevenwide]{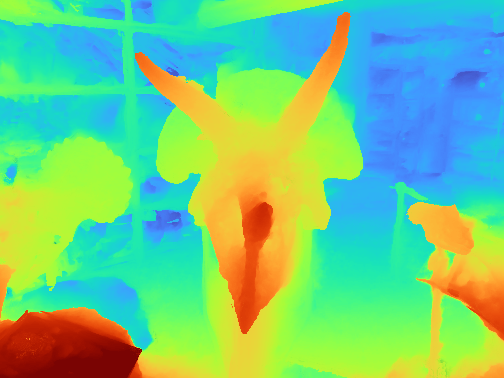} &
         \\
         \multirow{2}{*}[+1em]{\rotatebox{90}{ DTU (3)}}  &
         \includegraphics[width=\sevenwide]{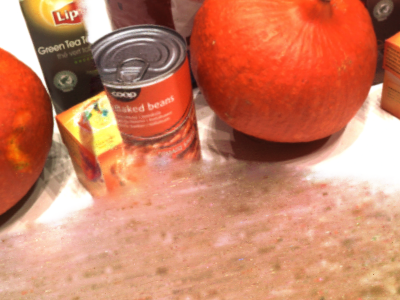} &
         \includegraphics[width=\sevenwide]{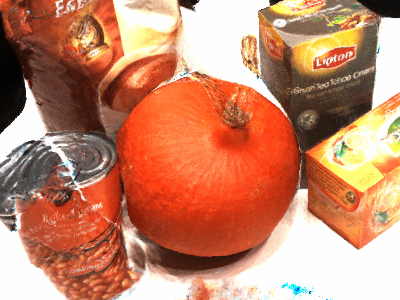} &
         \includegraphics[width=\sevenwide]{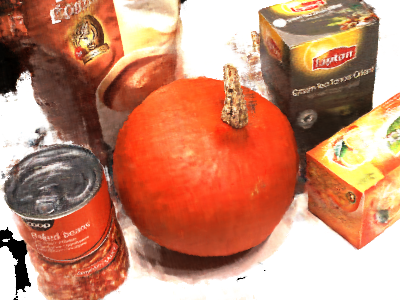} &
         \includegraphics[width=\sevenwide]{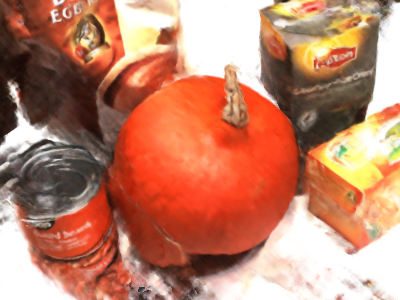} &
         \includegraphics[width=\sevenwide]{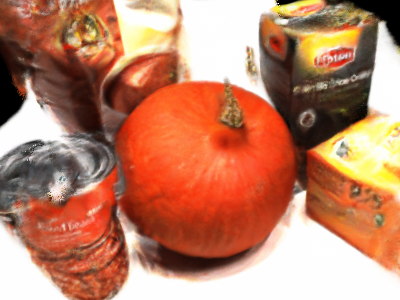} &
         \includegraphics[width=\sevenwide]{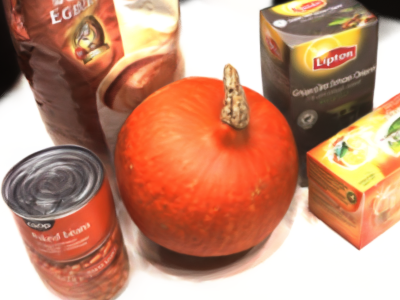} &
         \includegraphics[width=\sevenwide]{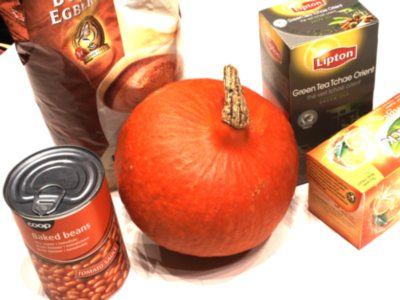} \\
          & %
         \includegraphics[width=\sevenwide]{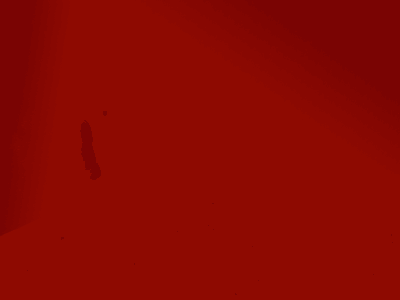} &
         \includegraphics[width=\sevenwide]{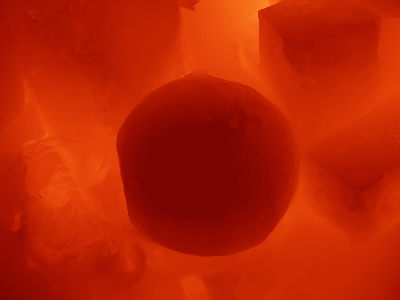} &
         \includegraphics[width=\sevenwide]{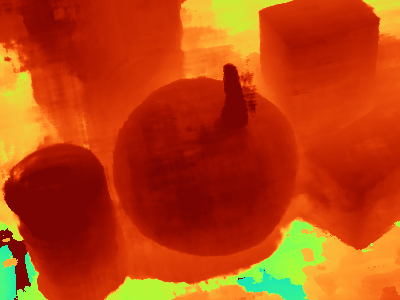} &
         \includegraphics[width=\sevenwide]{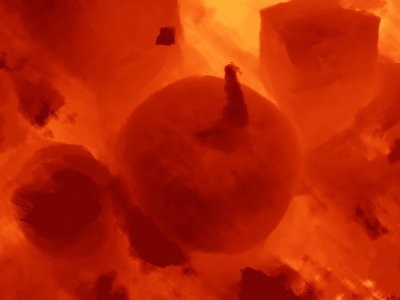} &
         \includegraphics[width=\sevenwide]{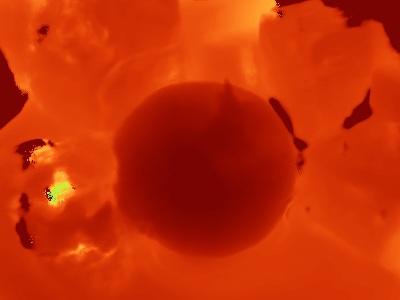} &
         \includegraphics[width=\sevenwide]{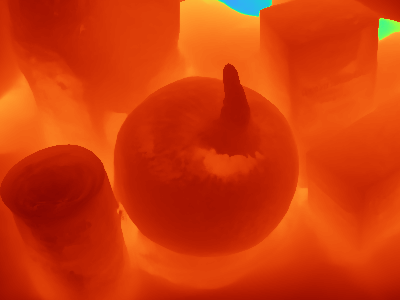} &
         \\

         \multirow{2}{*}[+1em]{\rotatebox{90}{ CO3D (6)}}  &
         \includegraphics[width=\sevenwide]{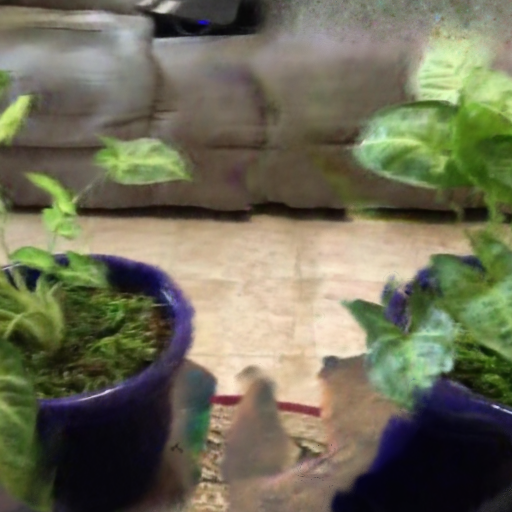} &
         \includegraphics[width=\sevenwide]{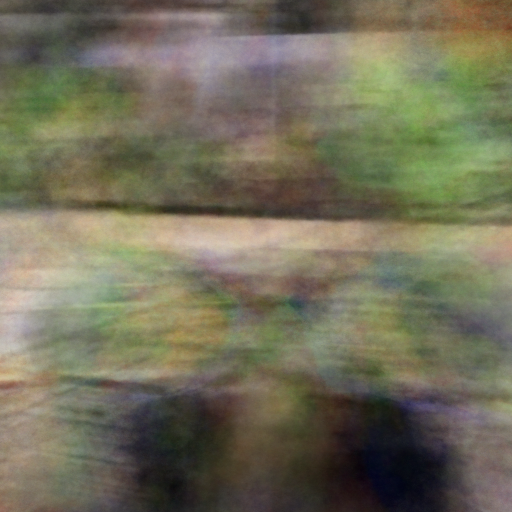} &
         \includegraphics[width=\sevenwide]{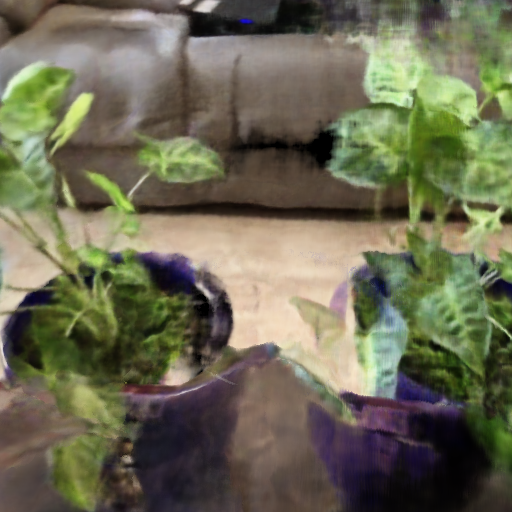} &
         \includegraphics[width=\sevenwide]{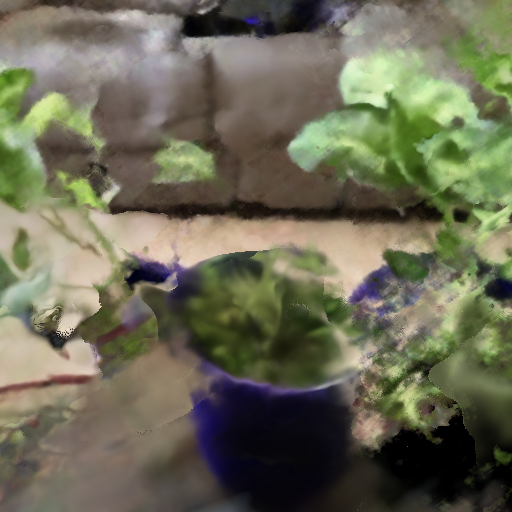} &
         \includegraphics[width=\sevenwide]{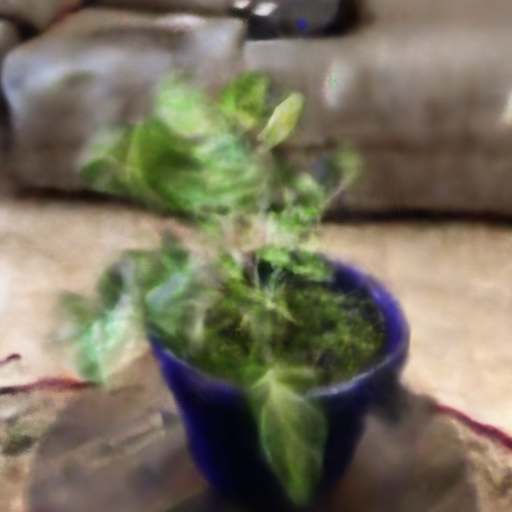} &
         \includegraphics[width=\sevenwide]{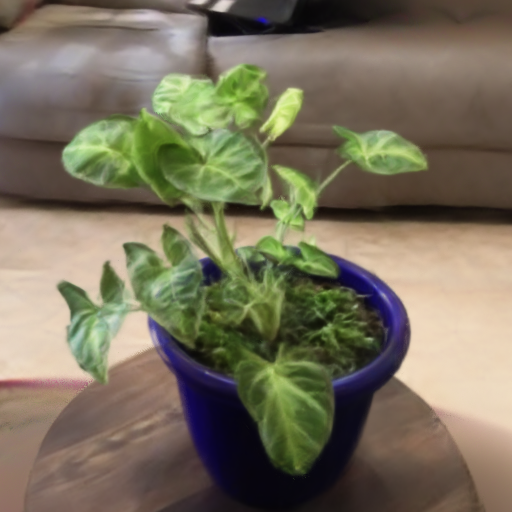} &
         \includegraphics[width=\sevenwide]{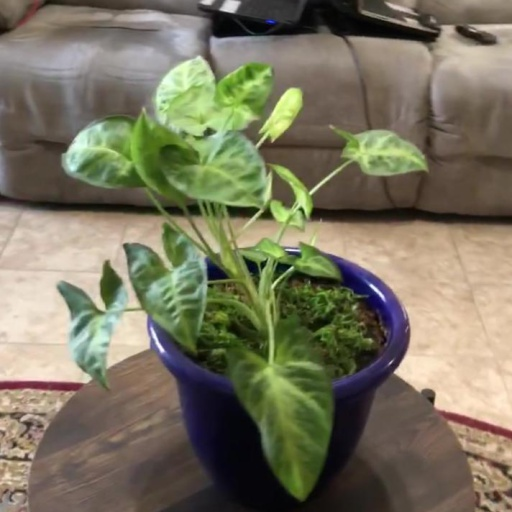} \\
          & %
         \includegraphics[width=\sevenwide]{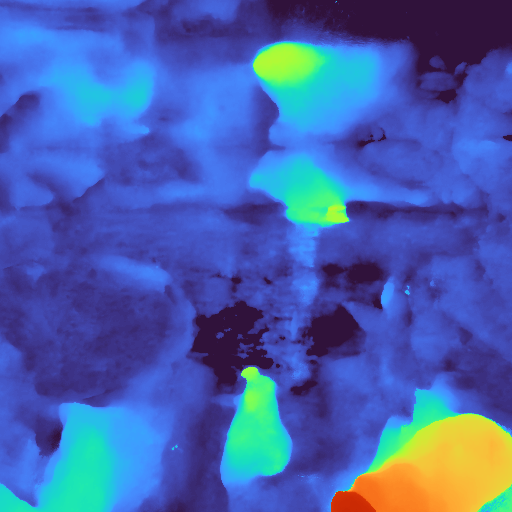} &
         \includegraphics[width=\sevenwide]{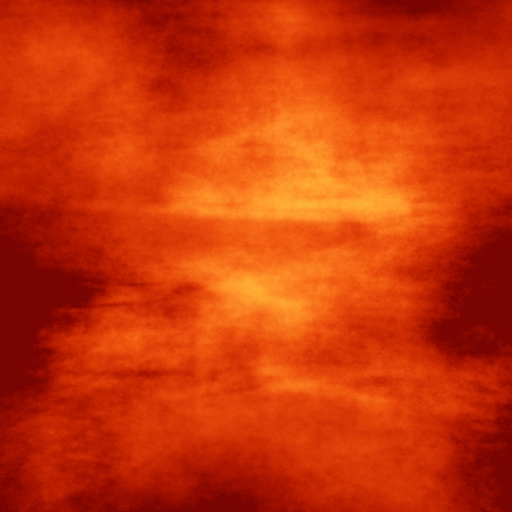} &
         \includegraphics[width=\sevenwide]{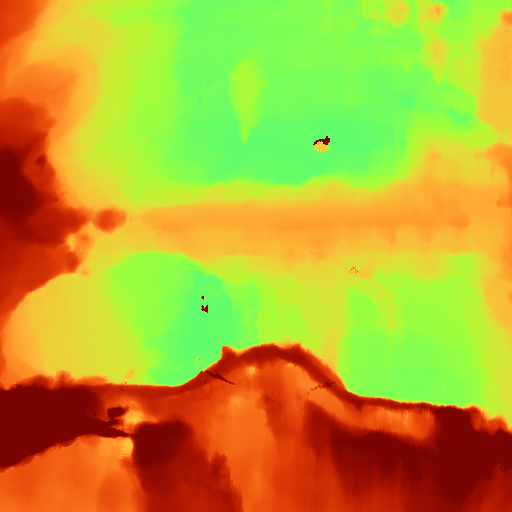} &
         \includegraphics[width=\sevenwide]{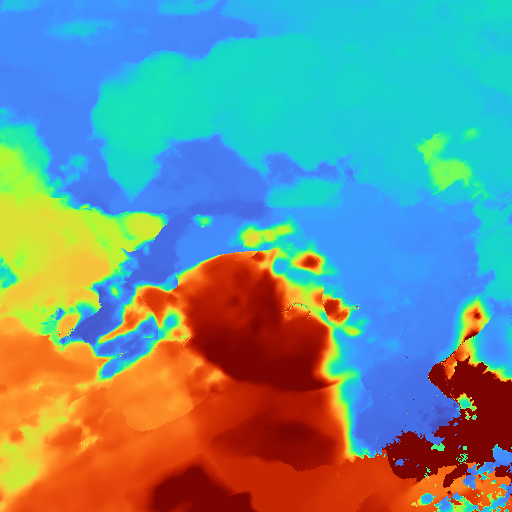} &
         \includegraphics[width=\sevenwide]{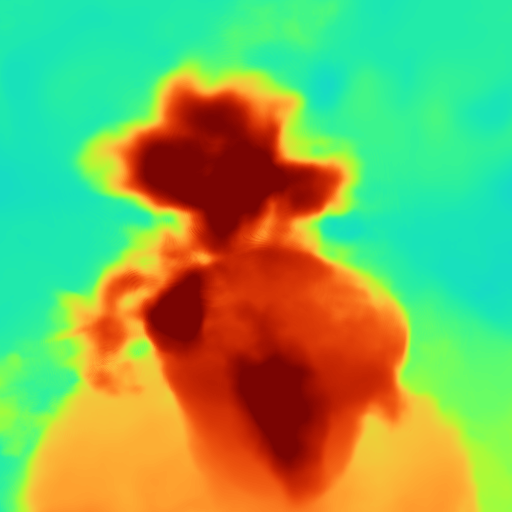} &
         \includegraphics[width=\sevenwide]{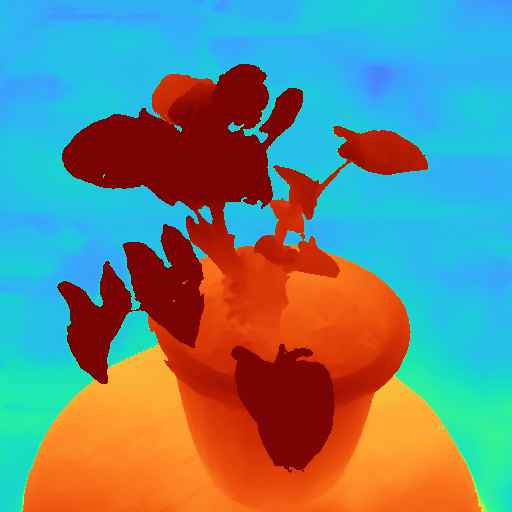} &
         \\
         \multirow{2}{*}[+2.5em]{\rotatebox[origin=c]{90}{ mip-NeRF 360 (9)}}  &
         \includegraphics[width=\sevenwide]{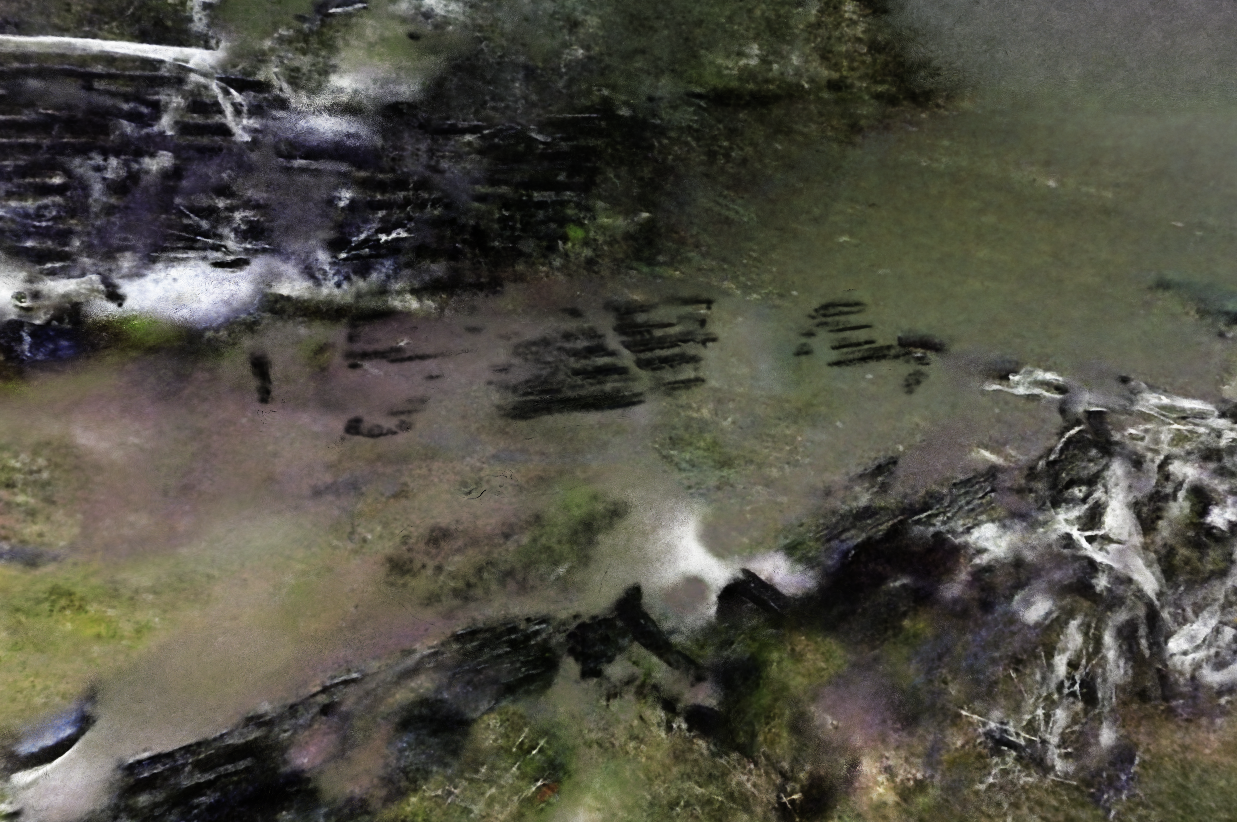} &
         \includegraphics[width=\sevenwide]{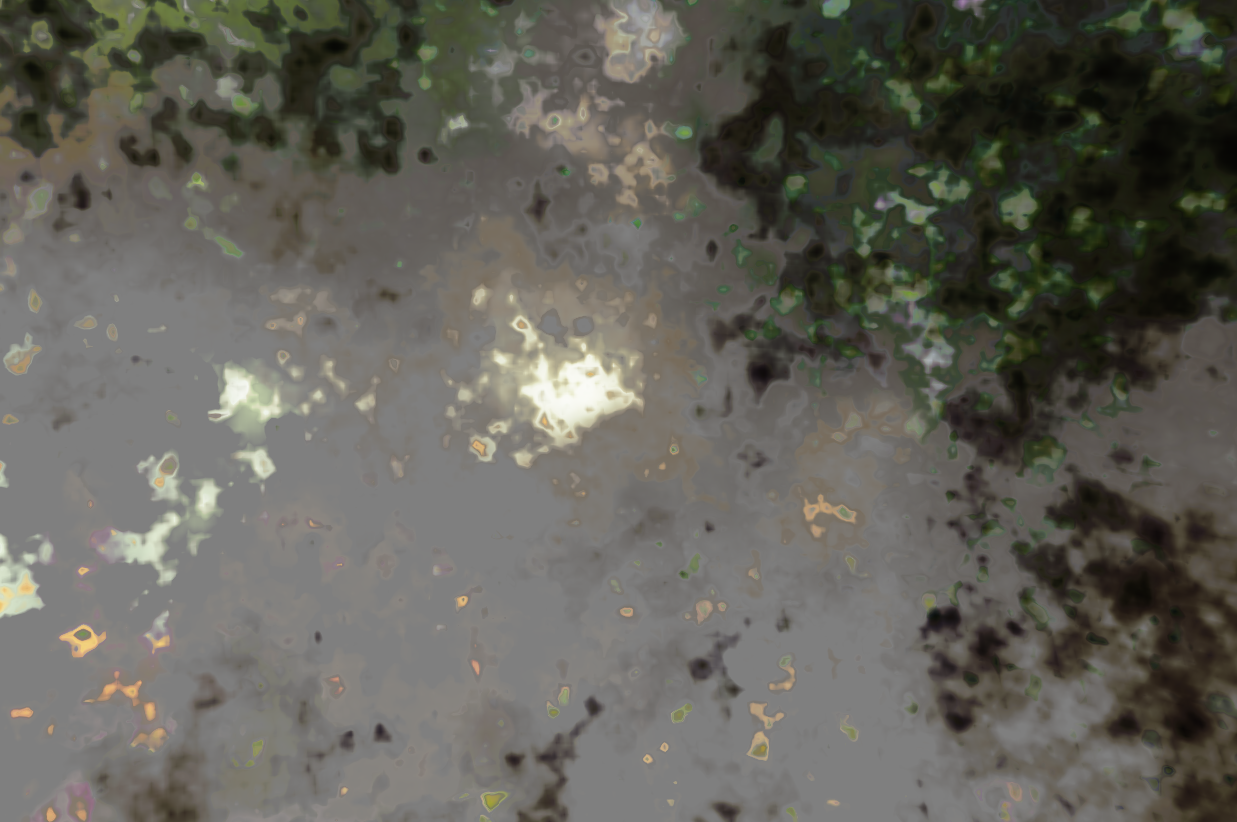} &
         \includegraphics[width=\sevenwide]{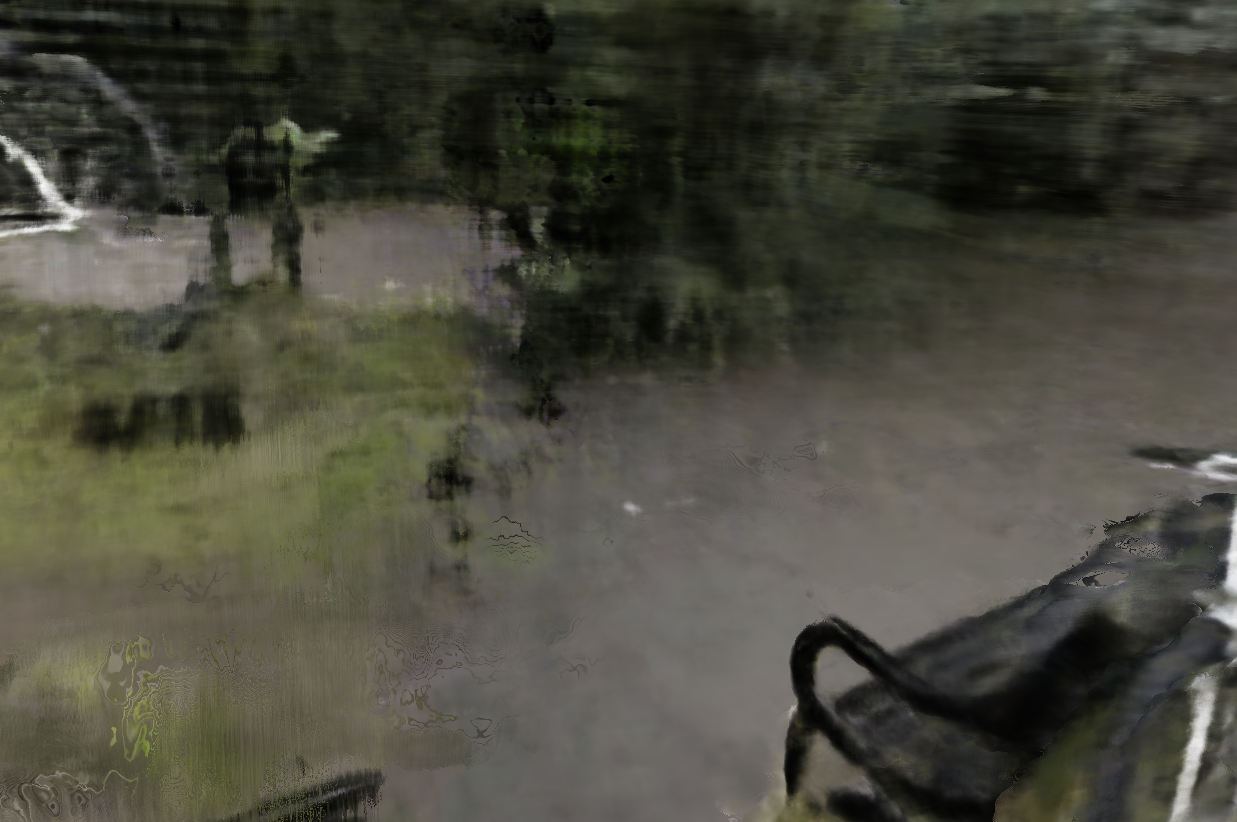} &
         \includegraphics[width=\sevenwide]{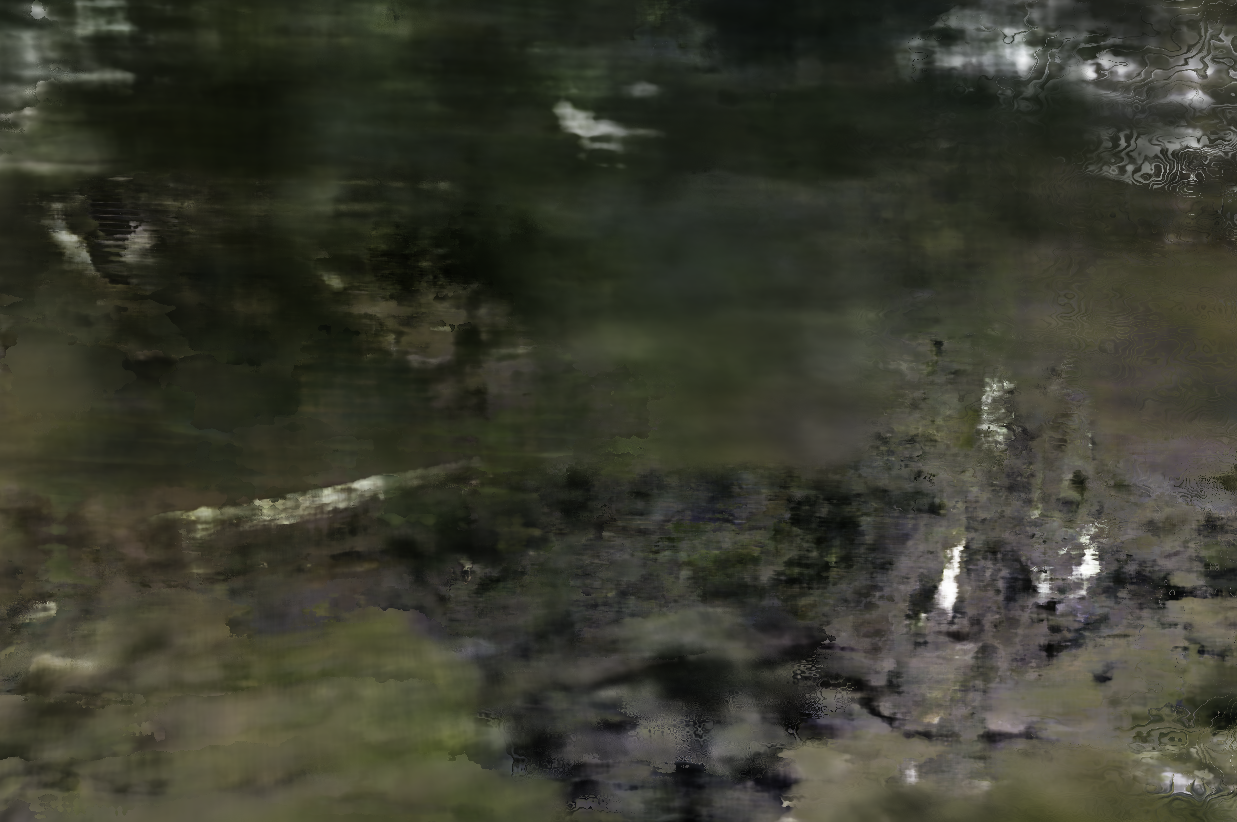} &
         \includegraphics[width=\sevenwide]{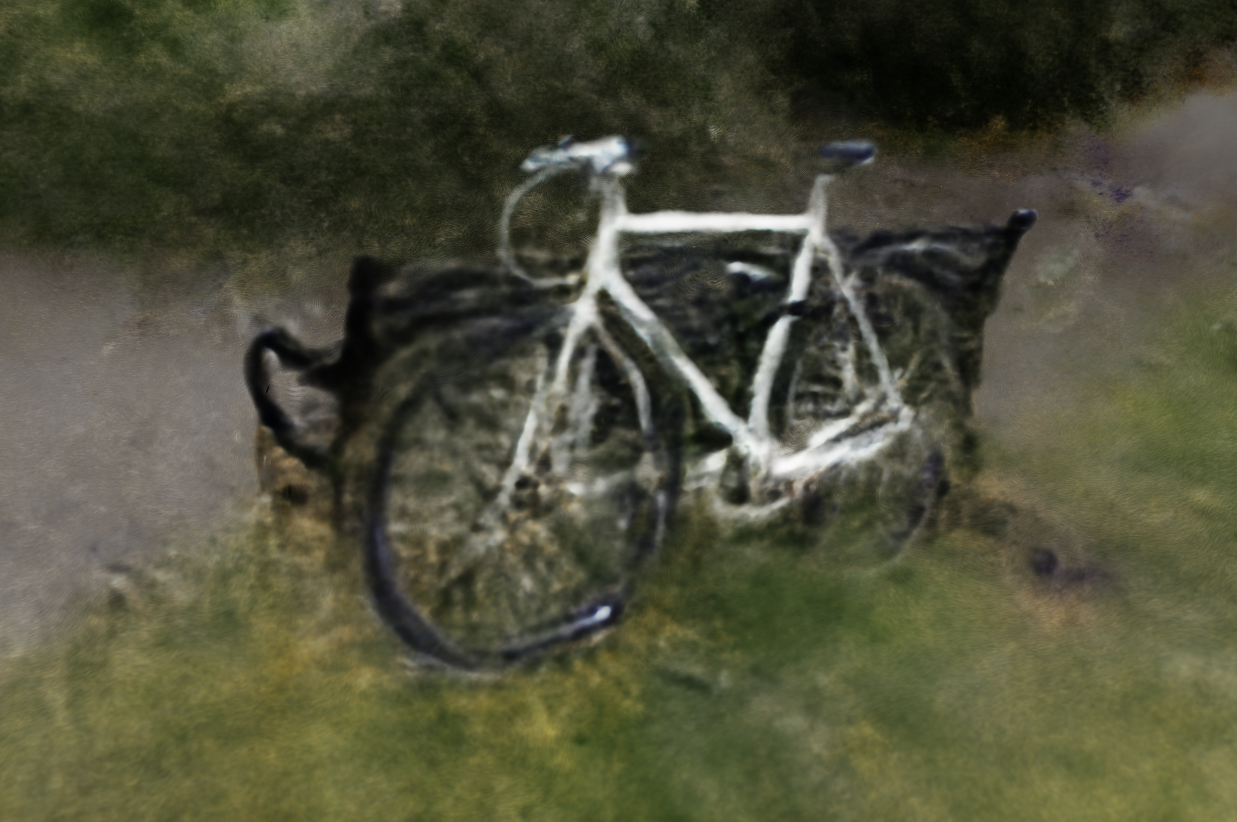} &
         \includegraphics[width=\sevenwide]{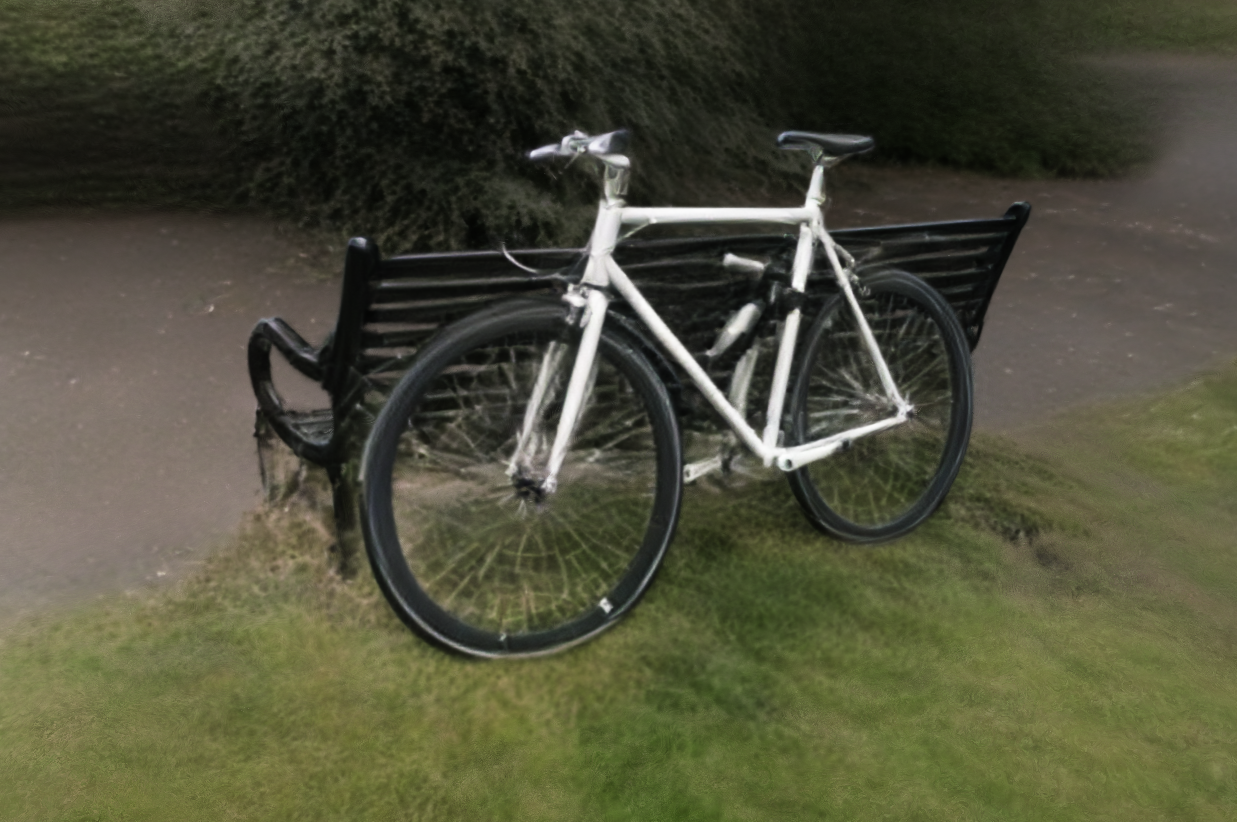} &
         \includegraphics[width=\sevenwide]{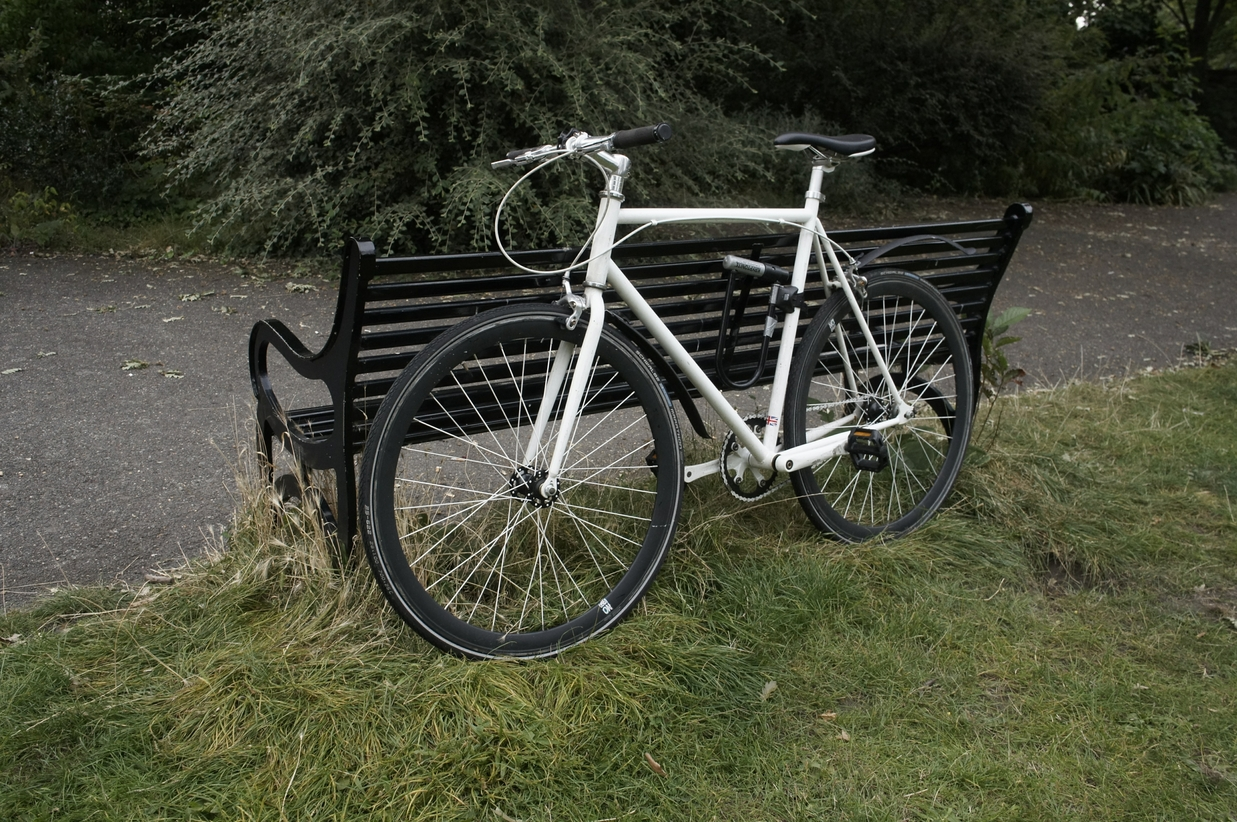} \\
          & %
         \includegraphics[width=\sevenwide]{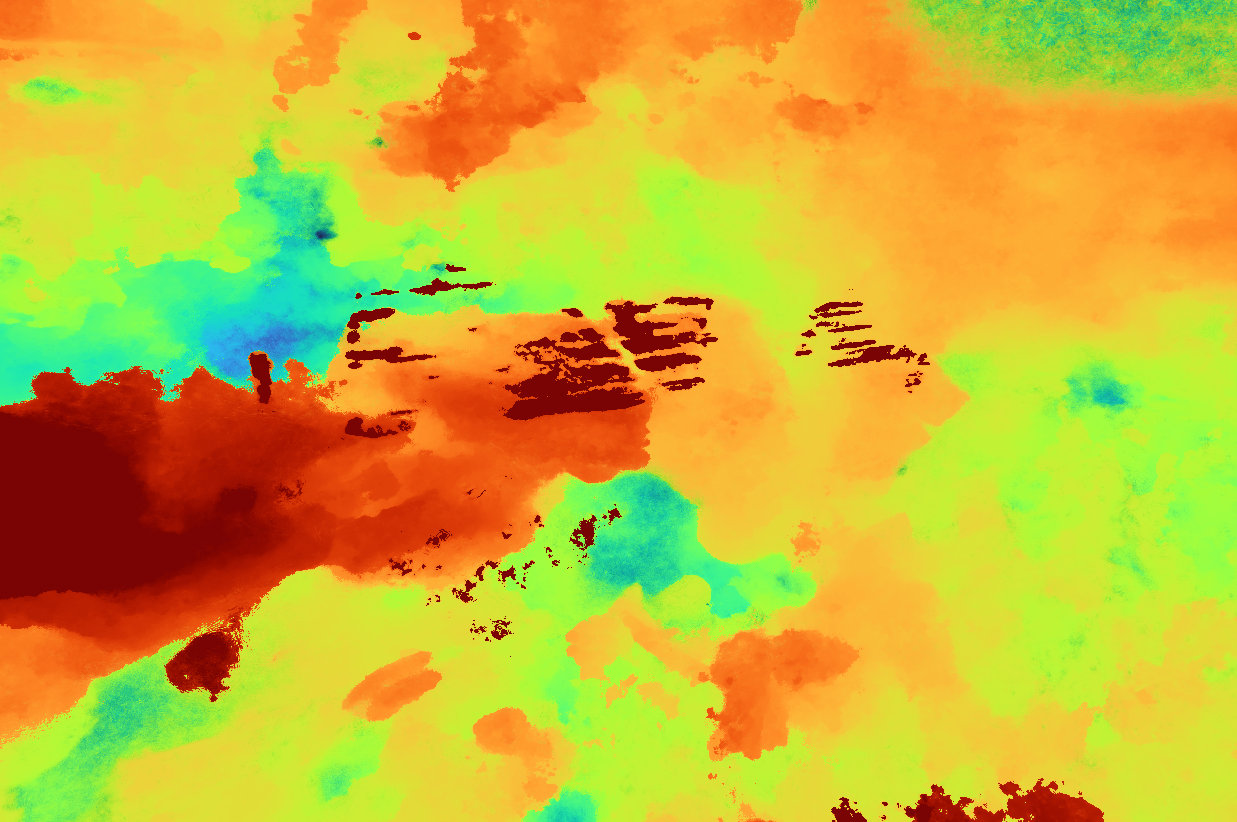} &
         \includegraphics[width=\sevenwide]{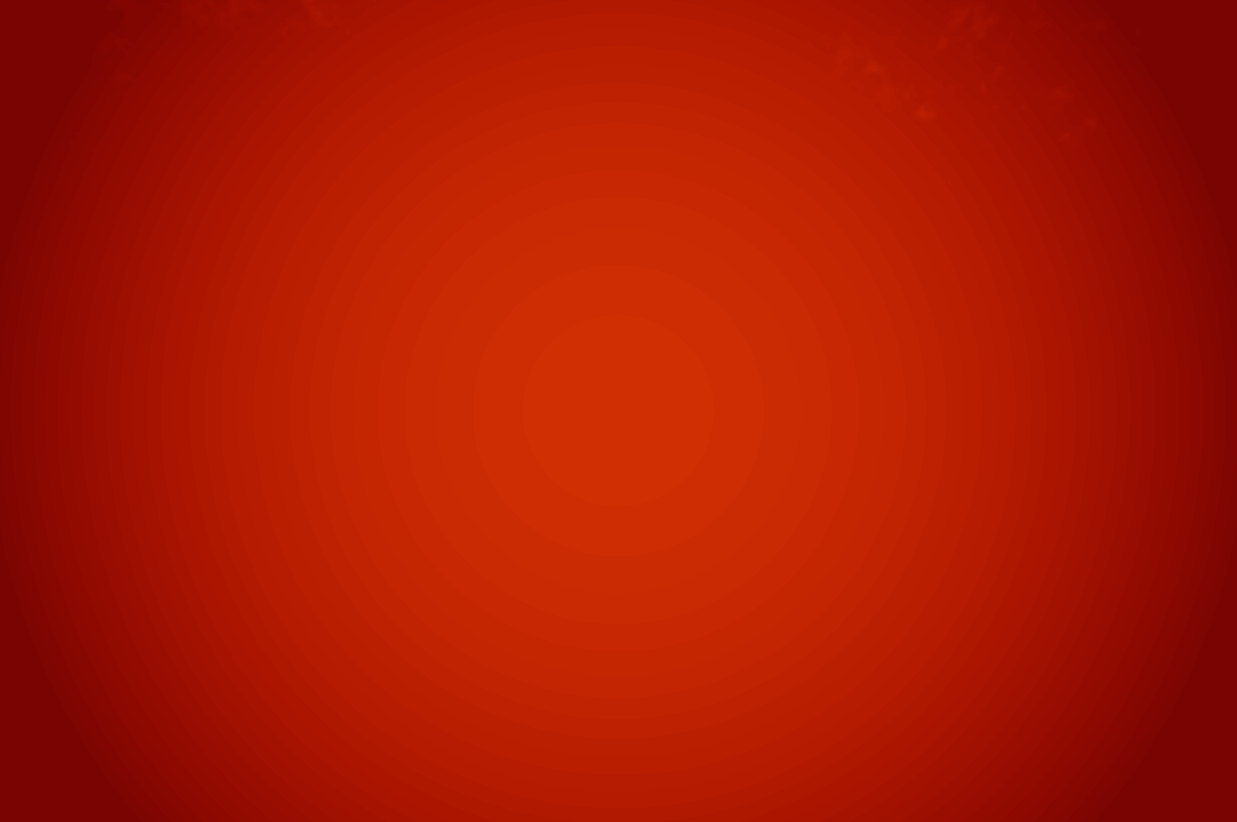} &
         \includegraphics[width=\sevenwide]{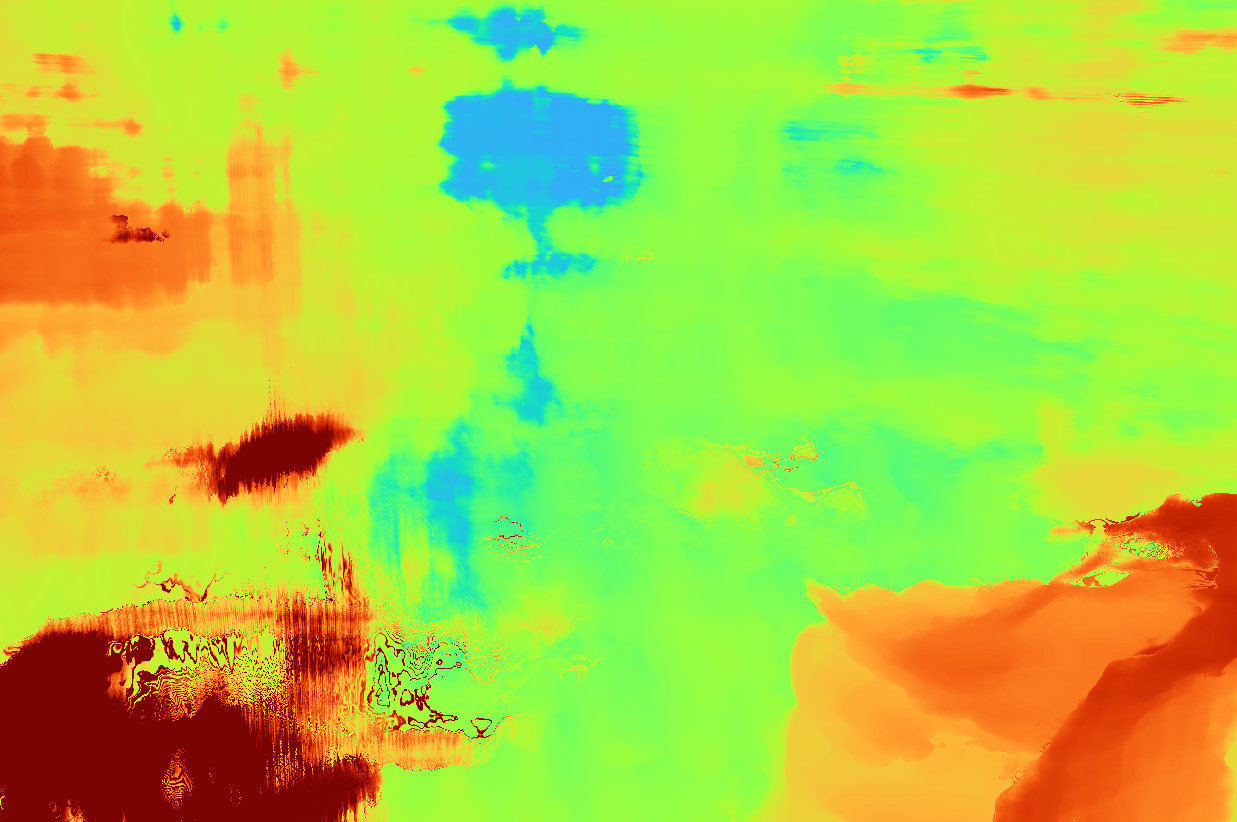} &
         \includegraphics[width=\sevenwide]{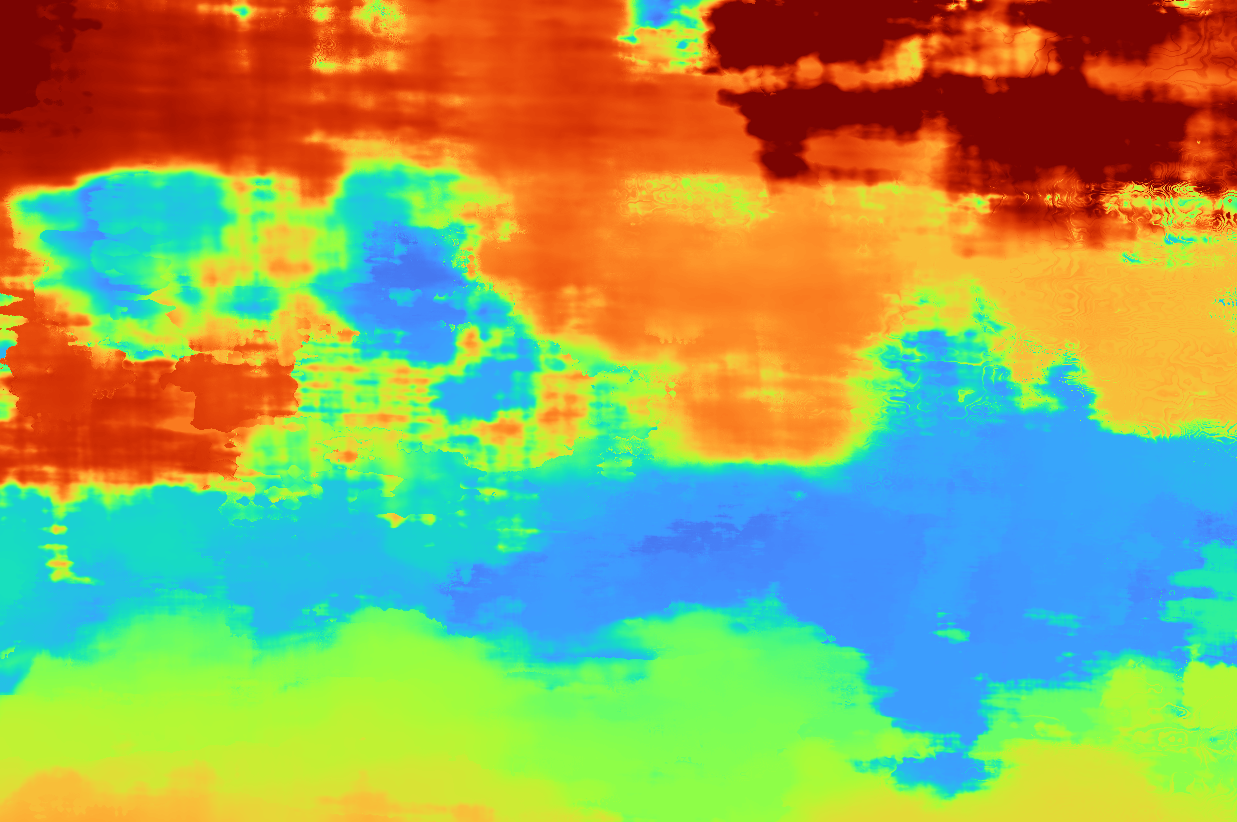} &
         \includegraphics[width=\sevenwide]{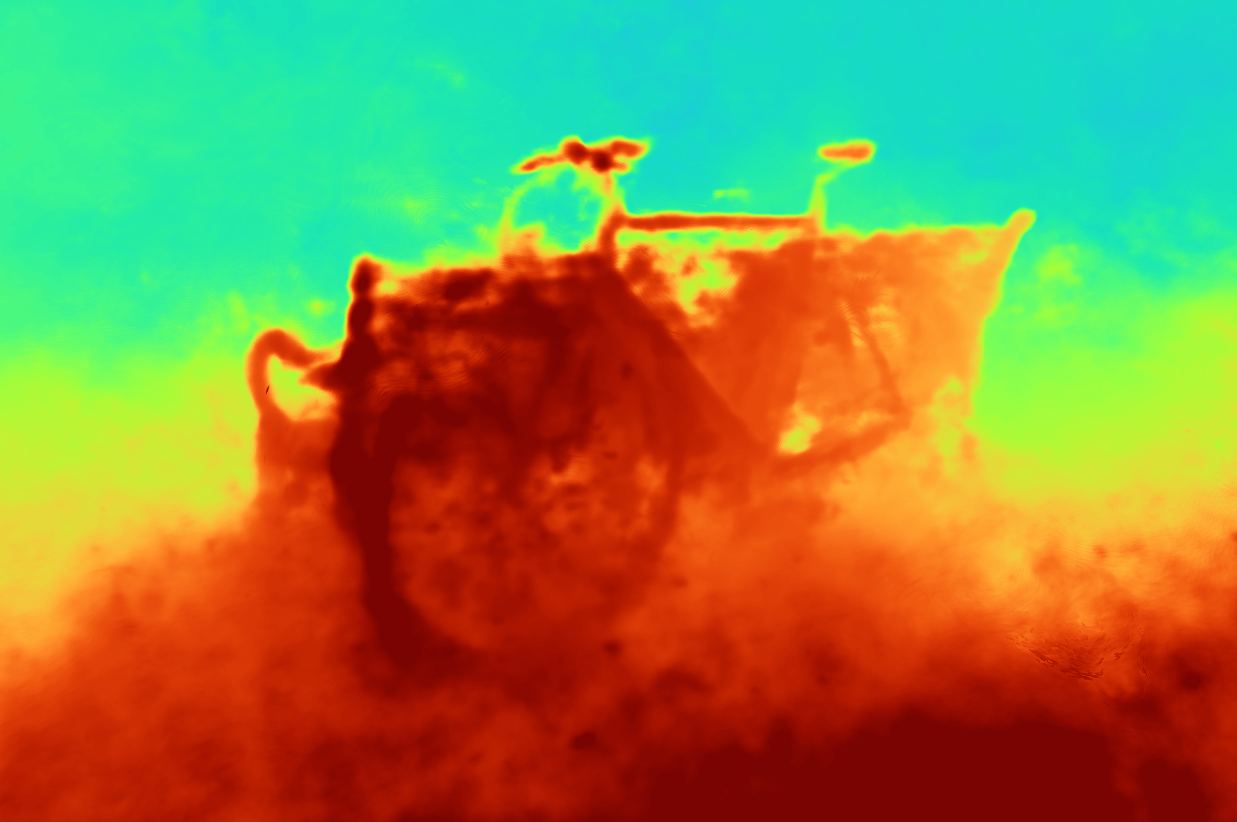} &
         \includegraphics[width=\sevenwide]{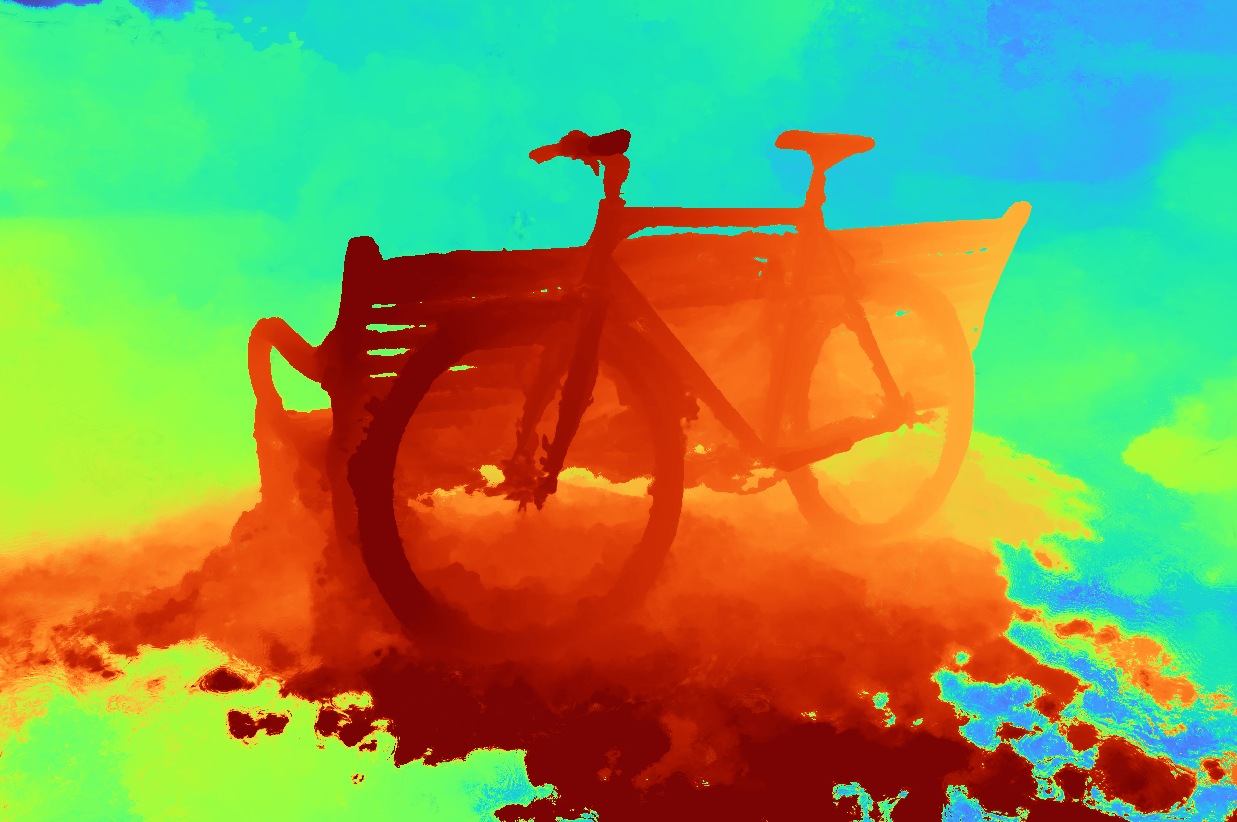} &
         \\
    \end{tabular}
    \caption{A visual comparison of rendered images and depth maps on scenes from the RealEstate10K~\cite{zhou2018stereo}, LLFF~\cite{mildenhall2019local}, DTU~\cite{jensen2014large}, CO3D~\cite{reizenstein2021common}, and mip-NeRF 360~\cite{barron2022mip} datasets (input view count indicated in parentheses). Both the appearance and geometry of our method are of higher quality than the baselines in these examples--typical failure modes exhibited by the baselines include ``floater'' artifacts visible in depth maps, color artifacts or blurry low-fidelity geometry in minimally observed regions of the scenes, correct texture appearing in incorrect locations in the image, and so on. We encourage the reader to watch our supplementary video, as many of these differences are easier to identify with a moving camera trajectory.}
    \label{fig:comparison2}
\end{figure*}

\subsection{Experiment Setup}
\myparagraph{Training Dataset}
To learn a generalizable diffusion prior for novel view synthesis, we train on a mixture of the synthetic Objaverse~\cite{deitke2023objaverse} dataset and three real-world datasets: CO3D~\cite{reizenstein2021common}, MVImgNet~\cite{yu2023mvimgnet}, and RealEstate10K~\cite{zhou2018stereo}. For Objaverse, we render each 3D asset from $16$ randomly sampled views at resolution $512\times 512$ and composite the rendering onto a randomly selected solid color background. For the other three real-world captured datasets, we center crop and resize each frame to $512\times 512$.
For training, we sample $3$ frames of the same scene as input views and sample another frame as the target view. 
Please refer to the supplementary materials for details about dataset mixing.

\myparagraph{Evaluation Dataset}
We evaluate our method on scenes from multiple datasets with 3, 6 and 9 input views, which include in-distribution datasets (CO3D~\cite{reizenstein2021common} and RealEstate10K~\cite{zhou2018stereo}) and out-of-distribution datasets (LLFF~\cite{mildenhall2019local}, DTU~\cite{jensen2014large} and mip-NeRF 360~\cite{barron2022mip}).
LLFF and DTU are datasets of forward-facing scenes, where we adhere to the evaluation protocol of RegNeRF~\cite{niemeyer2022regnerf}.
For the real-world object-centric scenes from CO3D we evaluate on a subset of $20$ scenes from $10$ categories.
RealEstate10K contains video clips gathered from YouTube, and we sample $10$ scenes (each with $100$ frames) from its test set for evaluation.
The mip-NeRF 360 dataset has $9$ indoor or outdoor scenes each containing a complex central object or area and a detailed background.
For CO3D and RealEstate10K, we select the input views evenly from all the frames and use every 8th of the remaining frames for evaluation.
For the mip-NeRF 360 dataset, we retain its original test set and select the input views from the training set using a heuristic to encourage reasonable camera spacing and coverage of the central object (see supplement for details).

\begin{table*}[t]
\centering
\resizebox{0.85\textwidth}{!}{%
\large
\begin{tabular}{cc|l|ccc|ccc|ccc}
 &
& & \multicolumn{3}{c|}{PSNR $\uparrow$} & \multicolumn{3}{c|}{SSIM $\uparrow$} & \multicolumn{3}{c}{LPIPS $\downarrow$} \\
                      &&            Method             & 3-view  & 6-view & 9-view & 3-view  & 6-view & 9-view & 3-view  & 6-view  & 9-view \\ \cmidrule{2-12}
\multirow{30}{*}{\rotatebox[origin=c]{90}{ {\huge$\underleftrightarrow{\large\hspace{20pt}\text{Harder}\hspace{140pt}\text{{Dataset Difficulty}}\hspace{140pt}\text{Easier}\hspace{20pt}}$} }} 
& \multirow{6}{*}{\rotatebox{90}{RealEstate10K}} 
 & \zipnerf     &  \cellcolor{tabthird}20.77 &  \cellcolor{tabthird}27.34 & \cellcolor{tabsecond}31.56 &  \cellcolor{tabthird}0.774 & \cellcolor{tabsecond}0.906 & \cellcolor{tabsecond}0.947 &  \cellcolor{tabthird}0.332 & \cellcolor{tabsecond}0.180 & \cellcolor{tabsecond}0.118 \\
&  & DiffusioNeRF &                      19.12 &                      24.18 &                      27.78 &                      0.710 &                      0.808 &                      0.869 &                      0.444 &                      0.344 &                      0.282 \\
&  & FreeNeRF     &                      20.54 &                      25.63 &                      27.32 &                      0.731 &                      0.817 &                      0.843 &                      0.394 &                      0.344 &                      0.332 \\
&  & SimpleNeRF   & \cellcolor{tabsecond}23.89 & \cellcolor{tabsecond}28.75 &  \cellcolor{tabthird}29.55 & \cellcolor{tabsecond}0.839 &  \cellcolor{tabthird}0.896 &  \cellcolor{tabthird}0.900 & \cellcolor{tabsecond}0.292 &  \cellcolor{tabthird}0.239 &  \cellcolor{tabthird}0.236 \\
 & & \zeronvs    &                      19.11 &                      22.54 &                      23.73 &                      0.675 &                      0.744 &                      0.766 &                      0.422 &                      0.374 &                      0.358 \\
 && Ours         &  \cellcolor{tabfirst}25.84 &  \cellcolor{tabfirst}29.99 &  \cellcolor{tabfirst}31.82 &  \cellcolor{tabfirst}0.910 &  \cellcolor{tabfirst}0.951 &  \cellcolor{tabfirst}0.961 &  \cellcolor{tabfirst}0.144 &  \cellcolor{tabfirst}0.103 &  \cellcolor{tabfirst}0.092   \\ \cmidrule{2-12}
& \multirow{7}{*}{\rotatebox{90}{LLFF}}
 & \zipnerf     &                      17.23 &                      20.71 &                      23.63 &                      0.574 &                      0.764 & \cellcolor{tabsecond}0.830 &                      0.373 & \cellcolor{tabsecond}0.221 & \cellcolor{tabsecond}0.166 \\
& & RegNeRF      &                      19.08 &                      23.09 &  \cellcolor{tabthird}24.84 &                      0.587 &                      0.760 &  \cellcolor{tabthird}0.820 &                      0.374 &                      0.243 &                      0.196 \\
&& DiffusioNeRF & \cellcolor{tabsecond}20.13 &  \cellcolor{tabthird}23.60 &                      24.62 & \cellcolor{tabsecond}0.631 & \cellcolor{tabsecond}0.775 &                      0.807 & \cellcolor{tabsecond}0.344 &                      0.235 &                      0.216 \\
&& FreeNeRF     &  \cellcolor{tabthird}19.63 & \cellcolor{tabsecond}23.72 & \cellcolor{tabsecond}25.12 &                      0.613 &  \cellcolor{tabthird}0.773 &  \cellcolor{tabthird}0.820 &  \cellcolor{tabthird}0.347 &  \cellcolor{tabthird}0.232 &  \cellcolor{tabthird}0.193 \\
 && SimpleNeRF   &                      19.24 &                      23.05 &                      23.98 &  \cellcolor{tabthird}0.623 &                      0.737 &                      0.762 &                      0.375 &                      0.296 &                      0.286 \\
 && \zeronvs    &                      15.91 &                      18.39 &                      18.79 &                      0.359 &                      0.449 &                      0.470 &                      0.512 &                      0.438 &                      0.416 \\
 && Ours         &  \cellcolor{tabfirst}21.34 &  \cellcolor{tabfirst}24.25 &  \cellcolor{tabfirst}25.21 &  \cellcolor{tabfirst}0.724 &  \cellcolor{tabfirst}0.815 &  \cellcolor{tabfirst}0.848 &  \cellcolor{tabfirst}0.203 &  \cellcolor{tabfirst}0.152 &  \cellcolor{tabfirst}0.134
      \\ \cmidrule{2-12}                
&\multirow{7}{*}{\rotatebox{90}{DTU}}  
  & \zipnerf     &                      9.18 &                      8.84 &                      9.23 &                      0.601 &                      0.589 &                      0.592 &                      0.383 &                      0.370 &                      0.364 \\
& & RegNeRF      &  \cellcolor{tabthird}19.39 &  \cellcolor{tabthird}22.24 & \cellcolor{tabsecond}24.62 &  \cellcolor{tabthird}0.777 &  \cellcolor{tabthird}0.850 &                      0.886 &  \cellcolor{tabthird}0.203 &  \cellcolor{tabthird}0.135 &  \cellcolor{tabthird}0.106 \\
& & DiffusioNeRF &                      16.14 &                      20.12 &  \cellcolor{tabthird}24.31 &                      0.731 &                      0.834 &  \cellcolor{tabthird}0.888 &                      0.221 &                      0.150 &                      0.111 \\
& & FreeNeRF     & \cellcolor{tabsecond}20.46 & \cellcolor{tabsecond}23.48 &  \cellcolor{tabfirst}25.56 & \cellcolor{tabsecond}0.826 & \cellcolor{tabsecond}0.870 & \cellcolor{tabsecond}0.902 & \cellcolor{tabsecond}0.173 & \cellcolor{tabsecond}0.131 & \cellcolor{tabsecond}0.102 \\
& & SimpleNeRF   &                      16.25 &                      20.60 &                      22.75 &                      0.751 &                      0.828 &                      0.856 &                      0.249 &                      0.190 &                      0.176 \\
& & \zeronvs    &                      16.71 &                      17.70 &                      17.92 &                      0.716 &                      0.737 &                      0.745 &                      0.223 &                      0.205 &                      0.200 \\
& & Ours         &  \cellcolor{tabfirst}20.74 &  \cellcolor{tabfirst}23.62 & \cellcolor{tabsecond}24.62 &  \cellcolor{tabfirst}0.875 &  \cellcolor{tabfirst}0.904 &  \cellcolor{tabfirst}0.921 &  \cellcolor{tabfirst}0.124 &  \cellcolor{tabfirst}0.105 &  \cellcolor{tabfirst}0.094
\\ \cmidrule{2-12}
&\multirow{7}{*}{\rotatebox{90}{CO3D}} 
 & \zipnerf     &                      14.34 &                      14.48 &                      14.97 &                      0.496 &                      0.497 &                      0.514 &                      0.652 &                      0.617 &                      0.590 \\
 && DiffusioNeRF &                      15.65 &                      18.05 &                      19.69 &  \cellcolor{tabthird}0.575 &                      0.603 &                      0.631 &  \cellcolor{tabthird}0.597 &                      0.544 &  \cellcolor{tabthird}0.500 \\
 && FreeNeRF     &                      13.28 &                      15.20 &                      17.35 &                      0.461 &                      0.523 &                      0.575 &                      0.634 &                      0.596 &                      0.561 \\
 && SimpleNeRF   &                      15.40 &                      18.12 & \cellcolor{tabsecond}20.52 &                      0.553 &  \cellcolor{tabthird}0.622 & \cellcolor{tabsecond}0.672 &                      0.612 &  \cellcolor{tabthird}0.541 & \cellcolor{tabsecond}0.493 \\
 && SparseFusion &  \cellcolor{tabthird}16.76 &  \cellcolor{tabthird}18.77 &                      19.13 &                      0.561 &                      0.600 &                      0.604 &                      0.695 &                      0.653 &                      0.651 \\
 && \zeronvs    & \cellcolor{tabsecond}17.13 & \cellcolor{tabsecond}19.72 &  \cellcolor{tabthird}20.50 & \cellcolor{tabsecond}0.581 & \cellcolor{tabsecond}0.627 &  \cellcolor{tabthird}0.640 & \cellcolor{tabsecond}0.566 & \cellcolor{tabsecond}0.515 &  \cellcolor{tabthird}0.500 \\
 && Ours         &  \cellcolor{tabfirst}19.59 &  \cellcolor{tabfirst}21.84 &  \cellcolor{tabfirst}22.95 &  \cellcolor{tabfirst}0.662 &  \cellcolor{tabfirst}0.714 &  \cellcolor{tabfirst}0.736 &  \cellcolor{tabfirst}0.398 &  \cellcolor{tabfirst}0.342 &  \cellcolor{tabfirst}0.318   \\ \cmidrule{2-12}                      
&\multirow{6}{*}{\rotatebox{90}{mip-NeRF 360}} 
 & \zipnerf     &                      12.77 &                      13.61 &                      14.30 &                      0.271 &                      0.284 &                      0.312 &  \cellcolor{tabthird}0.705 & \cellcolor{tabsecond}0.663 & \cellcolor{tabsecond}0.633 \\
&& DiffusioNeRF &                      11.05 &                      12.55 &                      13.37 &                      0.189 &                      0.255 &                      0.267 &                      0.735 &  \cellcolor{tabthird}0.692 &                      0.680 \\
&& FreeNeRF     &                      12.87 &                      13.35 &                      14.59 &                      0.260 &                      0.283 &                      0.319 &                      0.715 &                      0.717 &                      0.695 \\
 && SimpleNeRF   &  \cellcolor{tabthird}13.27 &  \cellcolor{tabthird}13.67 &  \cellcolor{tabthird}15.15 &  \cellcolor{tabthird}0.283 &  \cellcolor{tabthird}0.312 & \cellcolor{tabsecond}0.354 &                      0.741 &                      0.721 &                      0.676 \\
 && \zeronvs    & \cellcolor{tabsecond}14.44 & \cellcolor{tabsecond}15.51 & \cellcolor{tabsecond}15.99 & \cellcolor{tabsecond}0.316 & \cellcolor{tabsecond}0.337 &  \cellcolor{tabthird}0.350 & \cellcolor{tabsecond}0.680 & \cellcolor{tabsecond}0.663 &  \cellcolor{tabthird}0.655 \\
 && Ours         &  \cellcolor{tabfirst}15.50 &  \cellcolor{tabfirst}16.93 &  \cellcolor{tabfirst}18.19 &  \cellcolor{tabfirst}0.358 &  \cellcolor{tabfirst}0.401 &  \cellcolor{tabfirst}0.432 &  \cellcolor{tabfirst}0.585 &  \cellcolor{tabfirst}0.544 &  \cellcolor{tabfirst}0.511    \\
\end{tabular}
}
\caption{Quantitative evaluation of few-view 3D reconstruction methods. Datasets are ordered in terms of sparsity from easier (novel views are close to observed views) to harder (novel views are far from observed views).
Our method is the first to be evaluated on such a wide range of few-view real datasets. Despite this generality, we outperform all baselines across all domains. 
Baselines that we additionally tuned for the task of few-view reconstruction are indicated with $^*$.}
\label{tab:main_table}
\end{table*}

\myparagraph{Baselines}
For evaluation datasets, we compare against the state-of-the-art dense-view NeRF model Zip-NeRF~\cite{barron2023zip} (which is also the reconstruction pipeline used in our model), and state-of-the-art few-view NeRF regularization methods including DiffusioNeRF~\cite{wynn2023diffusionerf}, FreeNeRF~\cite{yang2023freenerf}, and SimpleNeRF~\cite{somraj2023simplenerf}.
We also compare to ZeroNVS~\cite{sargent2023zeronvs}, concurrent work on novel view synthesis of scenes from a single image. To adapt it to multiview inputs, we use the input view closest to the sampled view as its input condition and denote this method as ZeroNVS$^*$.
On the CO3D dataset we additionally compare to SparseFusion~\cite{zhou2023sparsefusion}.
Following their setup, we train the SparseFusion model in a category-specific manner.
However, unlike the original implementation, which masks out the foreground object and sidesteps the difficulty of recovering the background of the scene, we use the whole unmasked image. 
Please refer to the supplement for more details about baselines.

\subsection{Comparison Results}
\label{sec:results}
We report the quantitative results in \tabref{main_table} and show qualitative comparisons in \figref{comparison2}. Please see the supplementary video for more visuals.
Our backbone NeRF model, Zip-NeRF, often overfits to the input views and exhibits artifacts like ``foggy'' geometry or ``floaters''.
State-of-the-art few-view NeRF regularization methods (DiffusioNeRF, FreeNeRF and SimpleNeRF) are able to significantly improve the baseline quality on forward-facing scenes like LLFF and DTU.
However, they fall short on 360-degree scenes (\eg the CO3D dataset), where a large portion of the scene is undersampled or even unobserved due to the a much larger relative disparity between input views.
On such scenes, ZeroNVS serves as a strong baseline as it often can reconstruct a complete 3D scene, but with limited visual fidelity.
Our method outperforms all baselines on both in-distribution and out-of-distribution datasets, achieving state-of-the-art performance for few-view NeRF reconstructions.

\subsection{Ablation Studies}
\label{sec:ablation}
In \tabref{ablation} and \figref{ablation}, we ablate two aspects of our diffusion model: the use of pretrained diffusion model weights (PT) and conditioning signal.
To ablate ``PT,'' we train the diffusion model from scratch.
To ablate conditioning, in the \texttt{pose} experiment we replace our PixelNeRF module with a conditioning mechanism similar to ZeroNVS~\cite{sargent2023zeronvs} (which itself extends Zero-1-to-3~\cite{liu2023zero}). This alternative simply concatenates the latents of all input images to the U-Net input, and represents the relative camera transforms as vectors (relative translation and rotation quaternion) which are concatenated with the CLIP embeddings.
Both variants produce sampled images of lower quality, which subsequently degrades the NeRF reconstruction.

In \figref{ablation_3d}, we ablate the choice of diffusion loss and find standard SDS results contain more artifacts, and the multi-step diffusion loss effectively mitigates these artifacts. Additionally, annealing $t_\mathrm{min}$ leads to more details.

\begin{table}[t]
\centering

\resizebox{\columnwidth}{!}{
\Large

\begin{tabular}{c|ll|c@{\,}c@{\,}c|c@{\,}c@{\,}c}
& & & \multicolumn{3}{c|}{NeRF renders} & \multicolumn{3}{c}{Diffusion samples} \\
       &      PT  & Condition & PSNR$\uparrow$ & SSIM$\uparrow$ & LPIPS$\downarrow$ & PSNR$\uparrow$ & SSIM$\uparrow$ & LPIPS$\downarrow$ \\ \midrule
\multirow{3}{*}{\rotatebox{90}{\resizebox{0.8\width}{!}{In-domain}}}
 & \checkmark & \texttt{pose}        &  \cellcolor{tabthird}20.57 &  \cellcolor{tabthird}0.749 &  \cellcolor{tabthird}0.367 &  \cellcolor{tabthird}15.11 &  \cellcolor{tabthird}0.546 &  \cellcolor{tabthird}0.484 \\
 & & \texttt{pixelnerf} & \cellcolor{tabsecond}25.15 & \cellcolor{tabsecond}0.815 & \cellcolor{tabsecond}0.246 & \cellcolor{tabsecond}22.40 & \cellcolor{tabsecond}0.723 & \cellcolor{tabsecond}0.314 \\
 & \checkmark & \texttt{pixelnerf}    &  \cellcolor{tabfirst}25.34 &  \cellcolor{tabfirst}0.823 &  \cellcolor{tabfirst}0.232 &  \cellcolor{tabfirst}24.05 &  \cellcolor{tabfirst}0.751 &  \cellcolor{tabfirst}0.281    
\\ \midrule

\multirow{3}{*}{\rotatebox{90}{\resizebox{0.8\width}{!}{Out-domain}}}
 & \checkmark & \texttt{pose}        &  \cellcolor{tabthird}17.28 &  \cellcolor{tabthird}0.521 &  \cellcolor{tabthird}0.458 &  \cellcolor{tabthird}12.18 &  \cellcolor{tabthird}0.244 &  \cellcolor{tabthird}0.599 \\
 & & \texttt{pixelnerf} & \cellcolor{tabsecond}19.82 & \cellcolor{tabsecond}0.580 & \cellcolor{tabsecond}0.383 & \cellcolor{tabsecond}16.46 & \cellcolor{tabsecond}0.420 & \cellcolor{tabsecond}0.452 \\
 & \checkmark & \texttt{pixelnerf}    &  \cellcolor{tabfirst}20.23 &  \cellcolor{tabfirst}0.596 &  \cellcolor{tabfirst}0.355 &  \cellcolor{tabfirst}17.44 &  \cellcolor{tabfirst}0.464 &  \cellcolor{tabfirst}0.411  
\end{tabular}
}

\caption{
We ablate two aspects of our model: pretrained diffusion weights (PT) and conditioning. For PT, we initialize the diffusion model weights from a pretrained text-to-image model. \texttt{pose} uses a pose conditioning similar to ZeroNVS~\cite{sargent2023zeronvs} while \texttt{pixelnerf} uses our PixelNeRF conditioning (\secref{diffusion}). We evaluate our model on both \emph{in-domain} datasets (RealEstate10K, CO3D) and \emph{out-of domain} datasets (LLFF, mip-NeRF 360).
}
\label{tab:ablation}
\end{table}

\newcommand{\threewide}{0.98in}
\newcommand{\fourwide}{0.7in}
\begin{figure}[t]
    \centering
    \begin{tabular}{@{}c@{\,\,}c@{\,\,}c@{\,\,}c@{\,\,}c@{}}
         
         \rotatebox{90}{\resizebox{0.8\width}{!}{\quad\quad  \texttt{pose}}}  &
         \includegraphics[width=\fourwide]{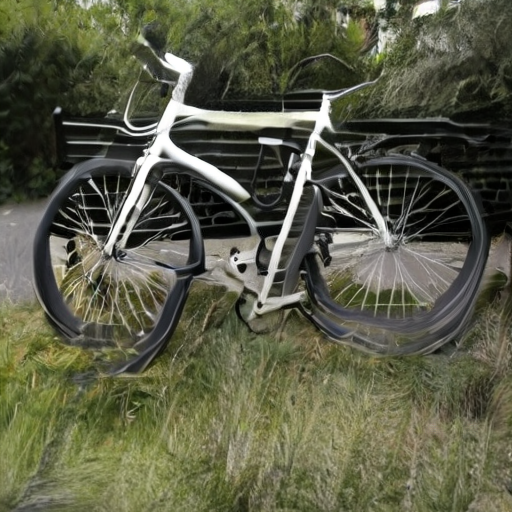} &
         \includegraphics[width=\fourwide]{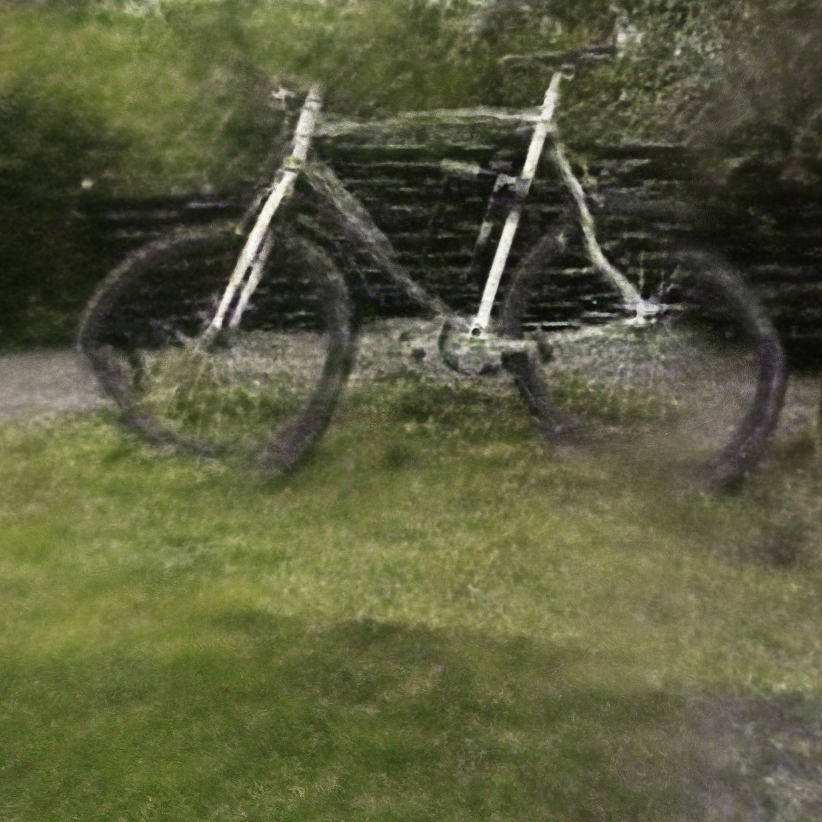} &
         \includegraphics[width=\fourwide]{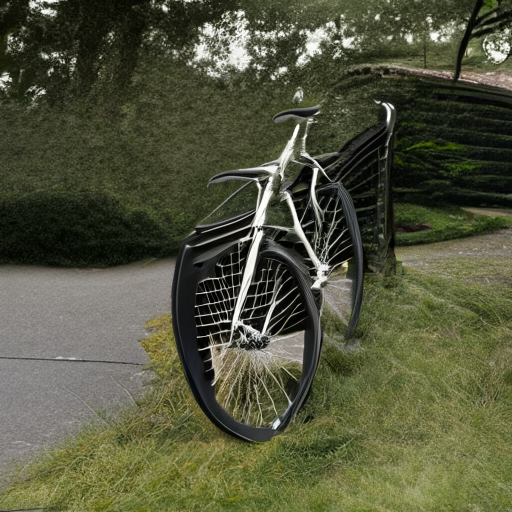} &
         \includegraphics[width=\fourwide]{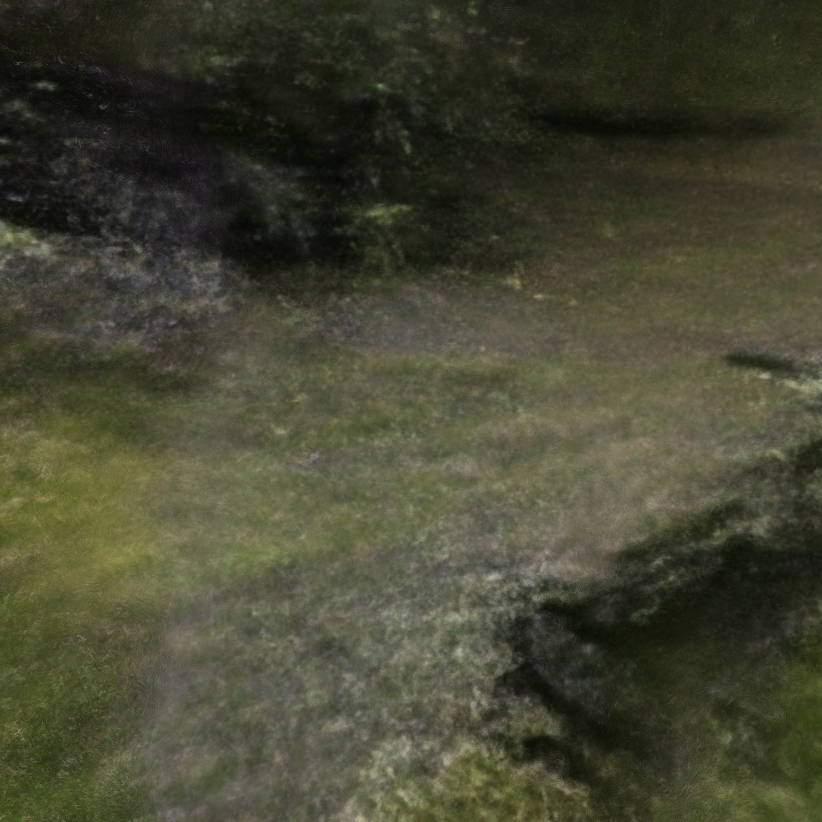} \\
         \rotatebox{90}{\resizebox{0.8\width}{!}{\quad\quad w/o PT}}  &
         \includegraphics[width=\fourwide]{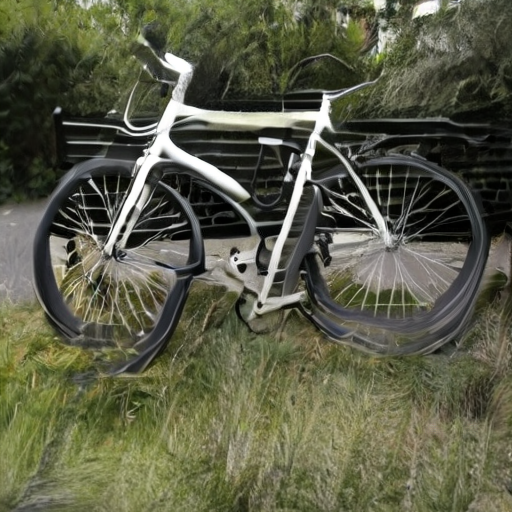} &
         \includegraphics[width=\fourwide]{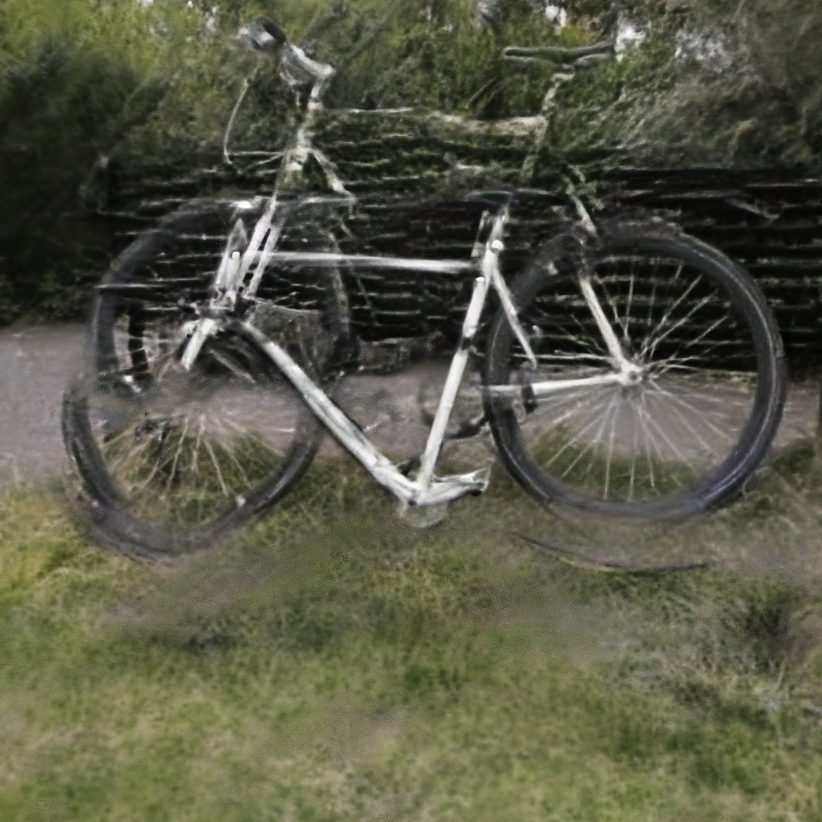} &
         \includegraphics[width=\fourwide]{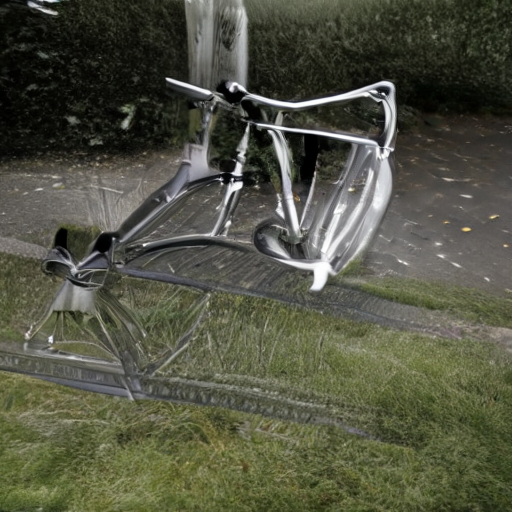} &
         \includegraphics[width=\fourwide]{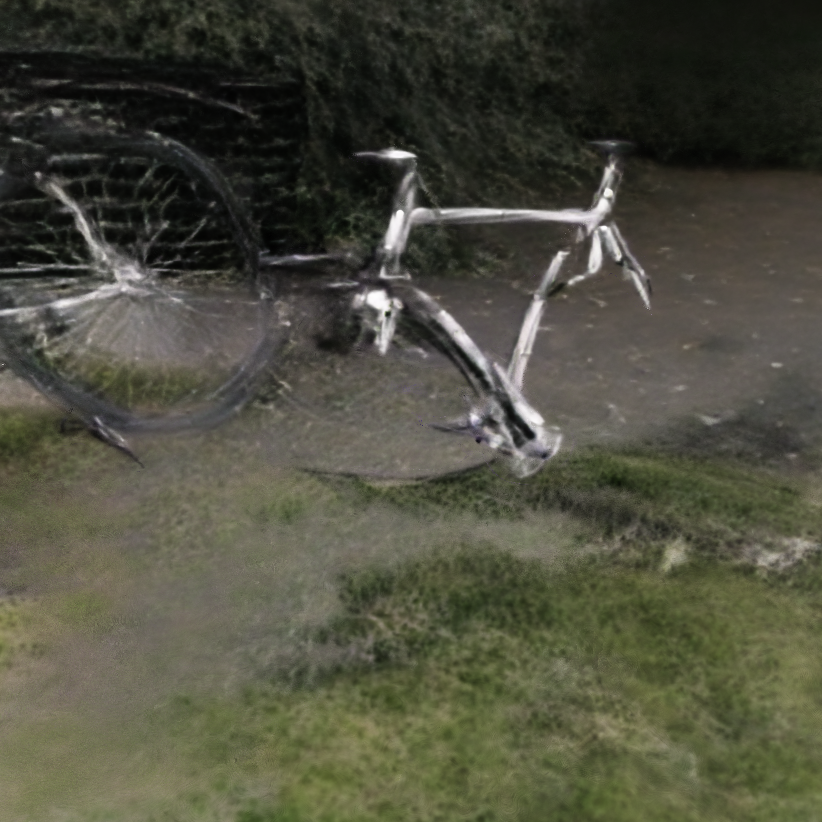} \\
         \rotatebox{90}{\resizebox{0.8\width}{!}{\quad\quad Ours}}  &
         \includegraphics[width=\fourwide]{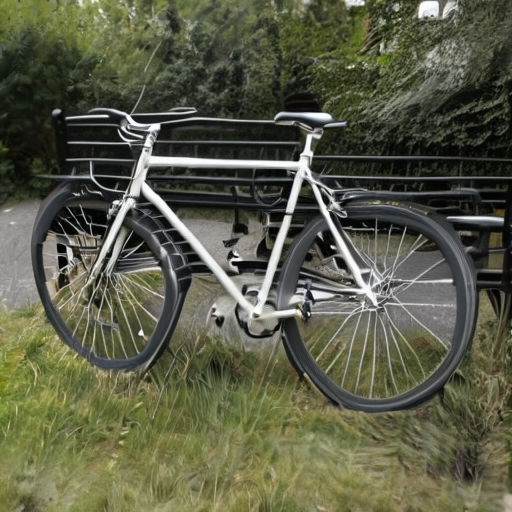} &
         \includegraphics[width=\fourwide]{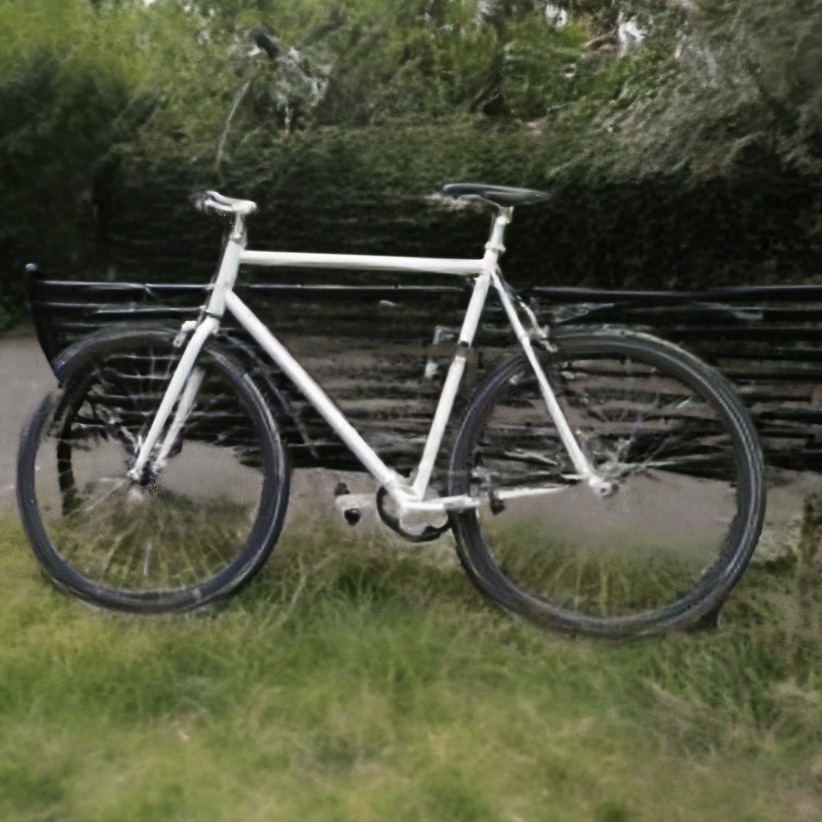} &
         \includegraphics[width=\fourwide]{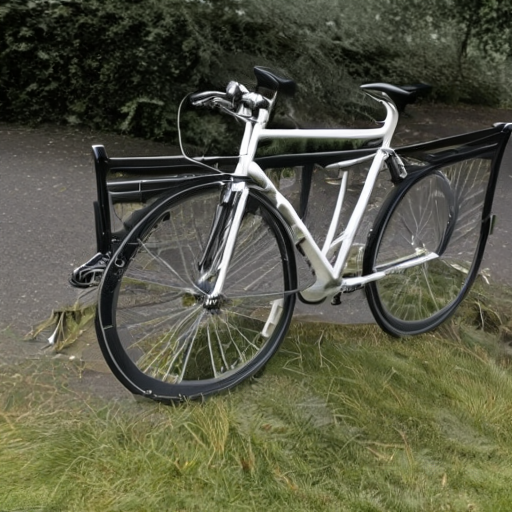} &
         \includegraphics[width=\fourwide]{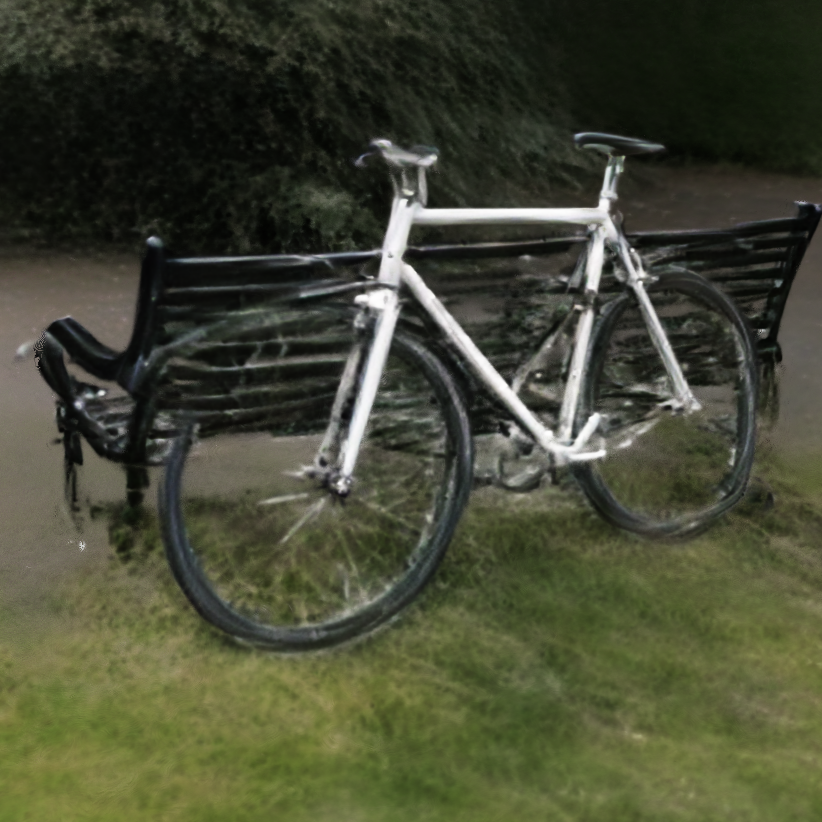} \\
         \rotatebox{90}{\resizebox{0.8\width}{!}{\, Ground Truth}}  &
         \includegraphics[width=\fourwide]{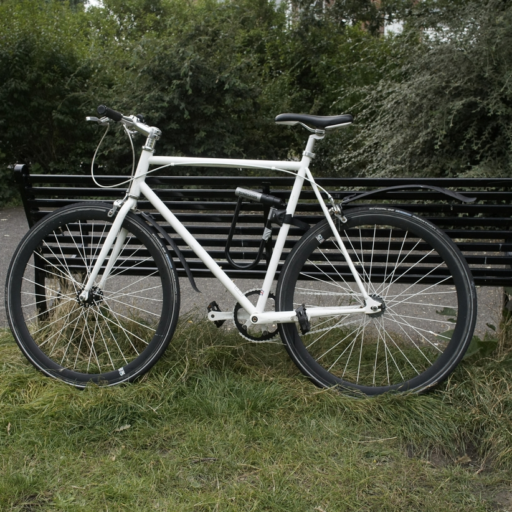} &
         \includegraphics[width=\fourwide]{figures/rgb_gt.png} &
         \includegraphics[width=\fourwide]{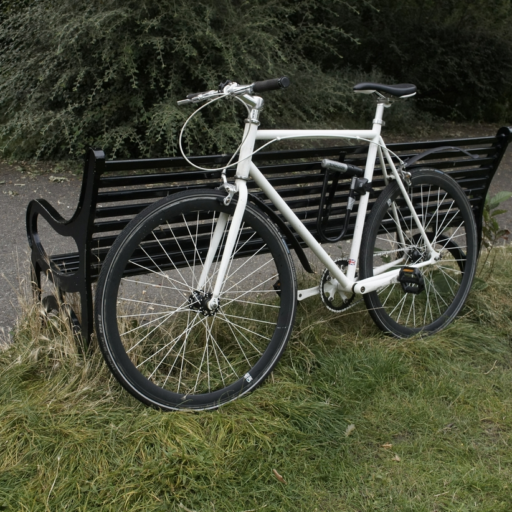} &
         \includegraphics[width=\fourwide]{figures/rgb_gt_2.png} \\
         & \resizebox{0.8\width}{!}{Sample 1} & \resizebox{0.8\width}{!}{NeRF render 1} & \resizebox{0.8\width}{!}{Sample 2} & \resizebox{0.8\width}{!}{NeRF render 2}  \\
    \end{tabular}
    \caption{\textbf{Ablation of diffusion model on 3-view reconstruction.} We show two samples from the diffusion models, and renderings from the reconstructed NeRFs under the same viewpoints for three variants of the diffusion model: \texttt{pose}, without pretraining, and our full model. The samples from nearby poses are inconsistent due to randomness, but can be successfully reconciled into an underlying NeRF reconstruction.}
    \label{fig:ablation}
\end{figure}

\begin{figure}[t]
    \centering
    \begin{tabular}{@{}c@{\,\,}c@{\,\,}c@{\,\,}c@{}}
         \rotatebox{90}{\,\,\,\,}  &
         \includegraphics[width=\threewide]{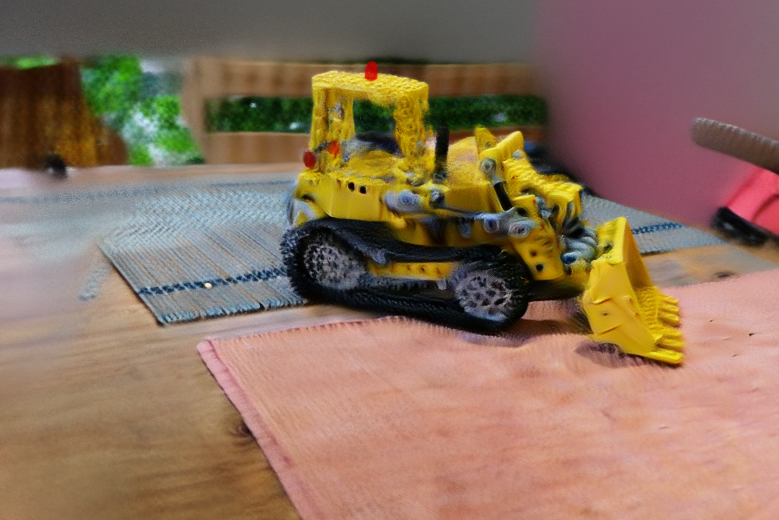} &
         \includegraphics[width=\threewide]{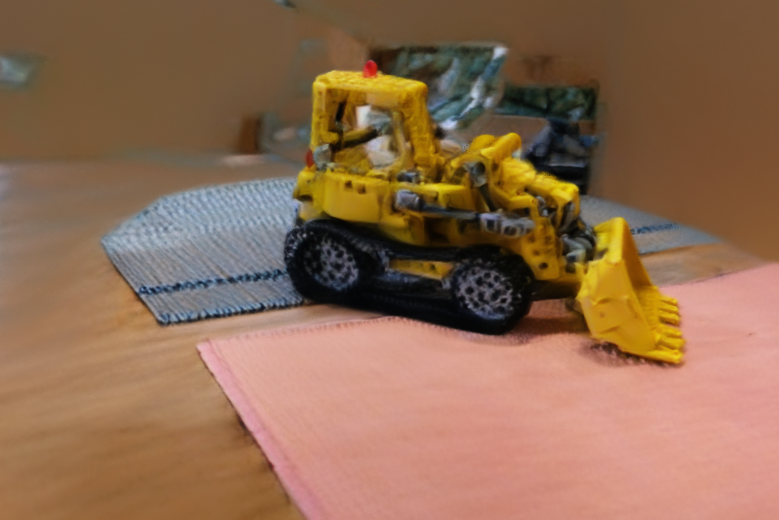} &
         \includegraphics[width=\threewide]{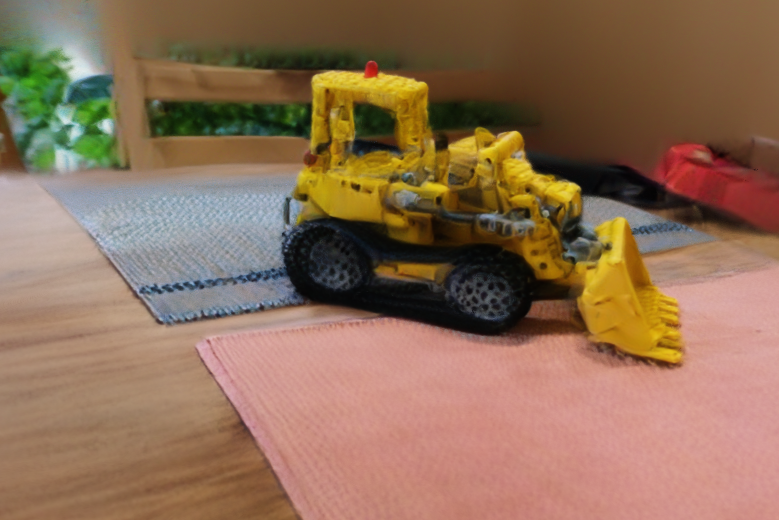} \\
         & \small{SDS} & \small{Multistep} & \small{Annealed Multistep} 
    \end{tabular}
    \caption{Comparing diffusion losses for 3D reconstruction. Note the ``blotchy'' texture on the placemat and background chair when using SDS, and improved background detail with annealing.}
    \label{fig:ablation_3d}
\end{figure}

\subsection{Scaling to More Views}
\label{sec:scaling}

\begin{figure}[t]
    \centering
    \includegraphics[trim={0.3cm 0.37cm 0.3cm 0.2cm},clip,width=\linewidth]{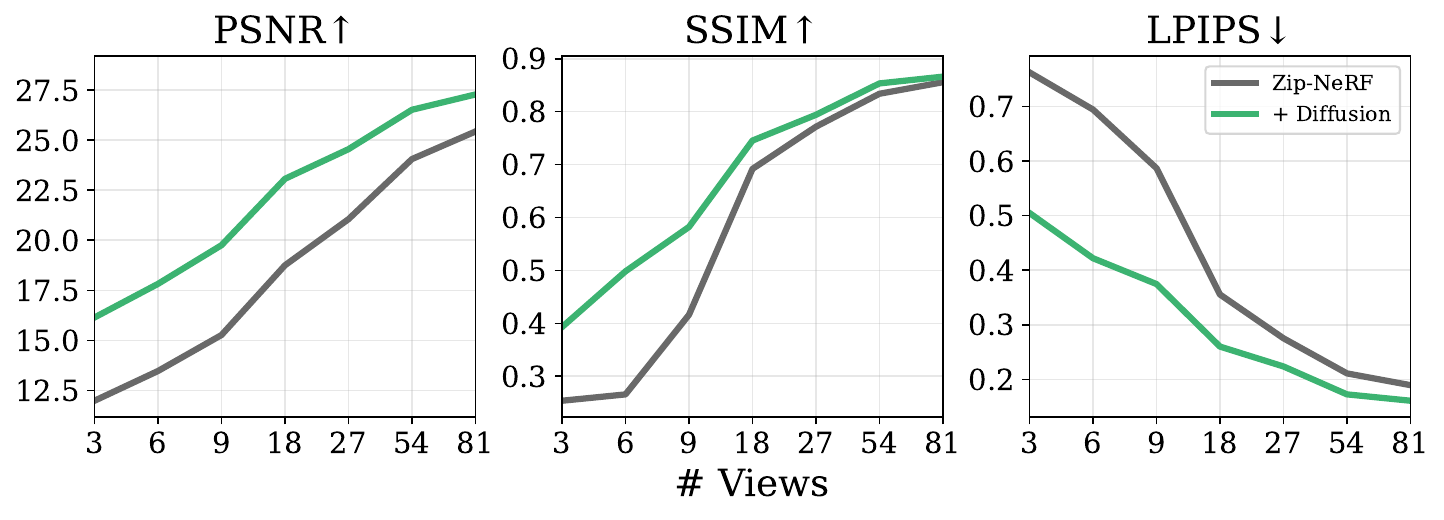}
    \vspace{-0.7cm}
    \caption{
    Our learned diffusion prior improves performance over the Zip-NeRF baseline
    up to as many as $81$ input views on the \texttt{kitchenlego} scene from the mip-NeRF 360 dataset. Though most results in this paper focus on the challenging case of 3-9 input views, achieving a high-quality reconstruction of a real-world scene often requires significantly more images--this plot demonstrates that our diffusion prior reduces this requirement at all points along the capture density vs. quality curve.}
    \label{fig:analysis_numbers}
\end{figure}

To further investigate the effectiveness and robustness of our diffusion prior, we evaluate our method and the backbone Zip-NeRF under various numbers of input views.
As the number of views increases, the input captures provide better coverage of the entire scene, resulting in less ambiguity.
Therefore, we set the weighting factor for our diffusion loss ($\mathcal{L}_{\mathrm{sample}}$) to be inversely proportional to the number of input views in this case.
As shown in \figref{analysis_numbers}, our test set performance is consistently better than Zip-NeRF, indicating that our diffusion prior can serve as an effective drop-in regularizer across a range of capture settings.

\section{Discussion}
\label{sec:conclusion}

The goal of \ours is to demonstrate the potential in piecing together two powerful building blocks. First, a state-of-the-art optimization-based 3D reconstruction pipeline, with an underlying 3D representation guaranteeing multiview consistency. And second, a powerful multiview-conditioned image diffusion model for generating plausible novel views, which can be used to guide reconstruction to avoid the artifacts resulting from an underconstrained inverse problem. 
Many current limitations are evident: the heavyweight diffusion model is costly and slows down reconstruction significantly; our current results demonstrate only limited 3D outpainting abilities compared to what our image model can hallucinate in 2D; tuning the balance of reconstruction and sample losses is tedious; etc.

However, this initial attempt at building such a system has already produced compelling results across a variety of scene types with significantly reduced view counts. We are optimistic that it may possibly serve as a template for improvements in sparse reconstruction, as we move toward a future of ever more accessible 3D reconstruction techniques with dramatically reduced capture requirements.

\paragraph{Acknowledgements}
We would like to thank Arthur Brussee, Ricardo Martin-Brualla, Rick Szeliski, Peter Hedman, and Jason Baldridge for their valuable contributions in discussing the project and reviewing the manuscript, and Zhicheng Wang for setting up some of the data loaders necessary for our diffusion model training pipeline. We are grateful to Randy Persaud and Henna Nandwani for infrastructure support.

{
    \small
    \bibliographystyle{ieeenat_fullname}
    \bibliography{main}
}

\setcounter{section}{0}
\renewcommand\thesection{\Alph{section}}

\section{Diffusion Model Details}
\label{sec:supdiff}

Our diffusion model is adapted from a pre-trained text-to-image latent diffusion model that maps $512 \times 512 \times 3$ inputs into a latent dimension of $64\times 64\times 8$. We modify this initial model to accept the necessary conditioning signals for text-free novel-view synthesis. We replace the inputs to the cross-attention pathway (which typically consist of a sequence of CLIP text embeddings) with the outputs of an additional dense layer. The input to this dense layer is a concatenated tensor consisting of (1) the unconditional CLIP text embedding (\ie the empty string \texttt{""}), and (2) the CLIP image embeddings of each of the input conditioning frames. We initialize the weights of this dense layer such that it produces the unconditional CLIP text embedding at the start of fine-tuning. This mechanism is inspired by the fine-tuning process in Zero-1-to-3~\cite{liu2023zero}. Zero-1-to-3 fine-tunes from an \emph{image variations} base model that has been previously fine-tuned to enable conditioning on CLIP image embeddings. We fine-tune directly from a text-conditioned model, but our architecture can learn image variation-like behavior through the dense layer. Furthermore, unlike Zero-1-to-3, our dense layer does not take pose as input, since the 3D transformation between the input conditioning frames and the target frame is applied through the PixelNeRF rendering process. In addition to the cross-attention modifications, we concatenate the outputs of the PixelNeRF model (a $64\times 64 \times 131$ tensor consisting of RGB and features) to the input noise that is passed to the U-Net. As in prior work~\cite{brooks2023instructpix2pix}, we initialize the additional convolutional weights to zero such that the added inputs have no effect at the start of fine-tuning. 

To enable classifier-free guidance on our added conditioning signals, we drop out all conditioning images for a training example with 10\% probability. We drop out the CLIP and PixelNeRF conditioning pathways independently in order to enable separate guidance, although we found empirically that using the same guidance weight across both conditioning signals (\ie performing joint CFG across both conditioning signals) produces optimal results. We train our model for $250,\!000$ iterations with a learning rate of $10^{-4}$ and a batch size of $128$. Our training examples consist of $3$ input conditioning images, $1$ target image, and the corresponding relative poses between each input image and the target image. This data is sampled from CO3D, RealEstate10k, MVImgNet, and Objaverse with uniform probability. For Objaverse, we light the object with random environment maps and compose it onto a random solid background at each training iteration.

\begin{figure*}
\centering
    \begin{subfigure}{0.245\linewidth}
        \includegraphics[width=\linewidth]{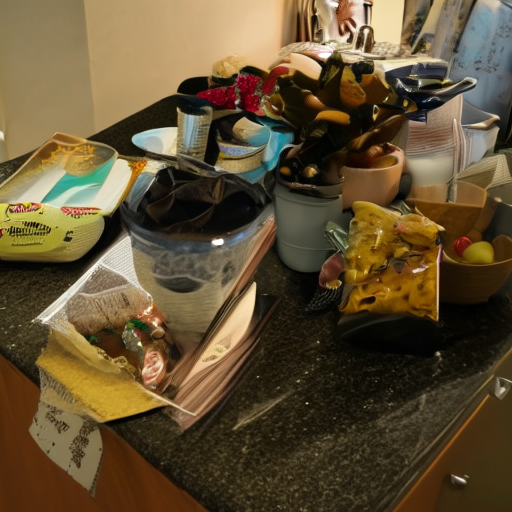}
        \caption{3 input views}
    \end{subfigure}
    \begin{subfigure}{0.245\linewidth}
        \includegraphics[width=\linewidth]{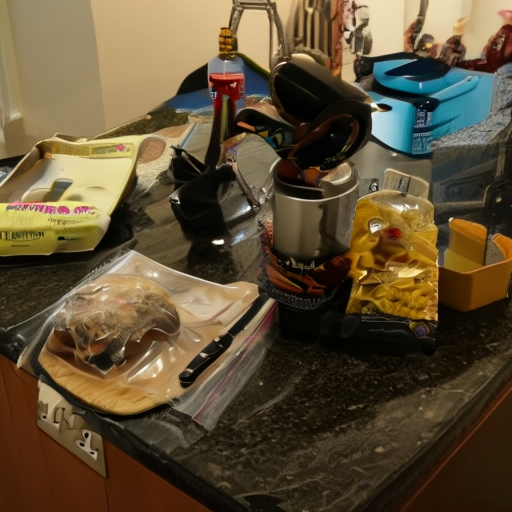}
        \caption{6 input views}
    \end{subfigure}
    \begin{subfigure}{0.245\linewidth}
        \includegraphics[width=\linewidth]{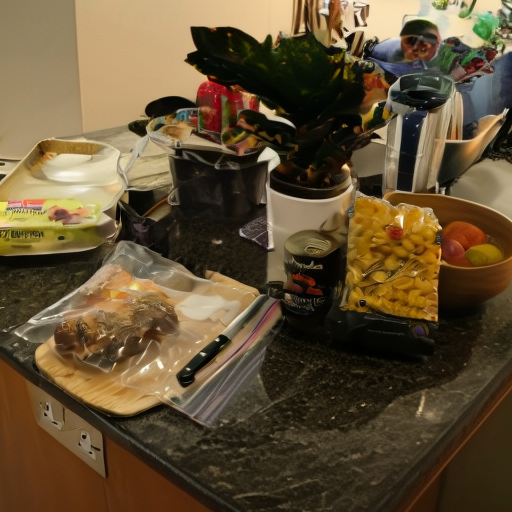}
        \caption{9 input views}
    \end{subfigure}
    \begin{subfigure}{0.245\linewidth}
        \includegraphics[width=\linewidth]{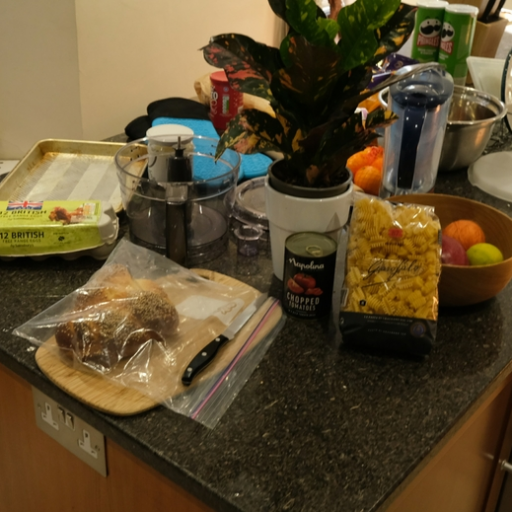}
        \caption{Ground truth}
    \end{subfigure}
    \caption{\textbf{More training views leads to better samples}. Here we show samples from the diffusion model  at a novel viewpoint while varying the number of training images. In all cases, the diffusion model is given the three nearest images to the novel viewpoint. We see that as we increase the number of known scene observations, the expected distance to the nearest training view decreases, therefore increasing the fidelity of diffusion model samples.}
\end{figure*}

\begin{figure*}
\centering
\begin{tabular}{@{}c@{}c@{\,}|@{\,}c@{}c@{}}
    \begin{subfigure}{0.245\linewidth}
        \includegraphics[width=\linewidth]{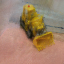}\\
        \includegraphics[width=\linewidth]{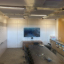}\\
        \includegraphics[width=\linewidth]{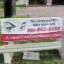}
        \caption{PixelNeRF RGB output}
    \end{subfigure} &
    \begin{subfigure}{0.245\linewidth}
        \includegraphics[width=\linewidth]{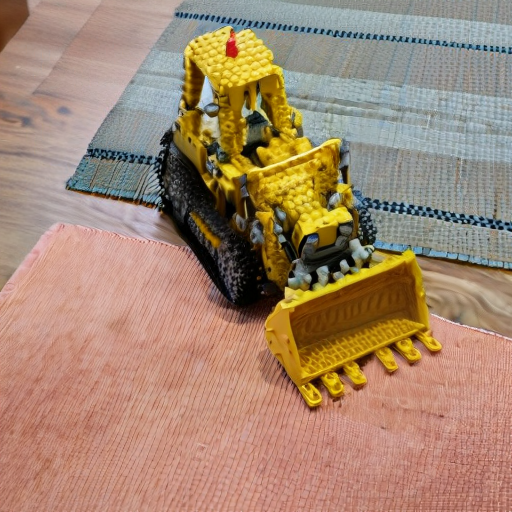}\\
        \includegraphics[width=\linewidth]{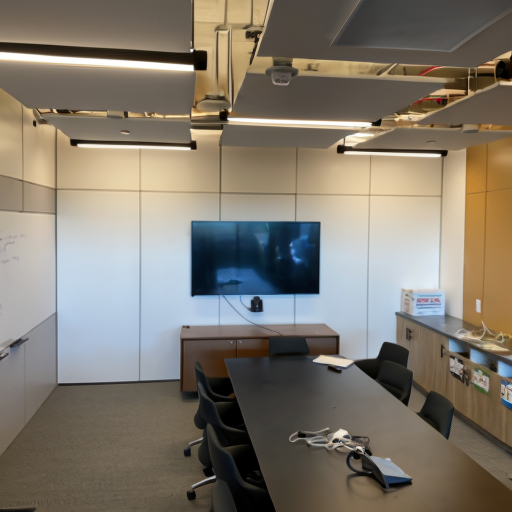}\\
        \includegraphics[width=\linewidth]{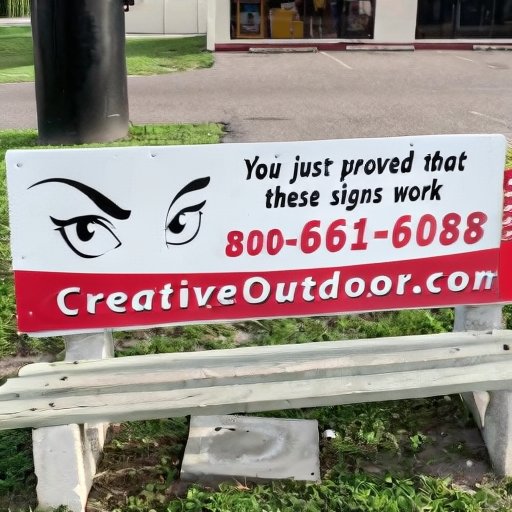}
        \caption{Diffusion model sample}
    \end{subfigure} &
    \begin{subfigure}{0.245\linewidth}
        \includegraphics[width=\linewidth]{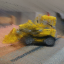}\\
        \includegraphics[width=\linewidth]{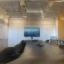}\\
        \includegraphics[width=\linewidth]{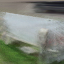}
        \caption{PixelNeRF RGB output}
    \end{subfigure} &
    \begin{subfigure}{0.245\linewidth}
        \includegraphics[width=\linewidth]{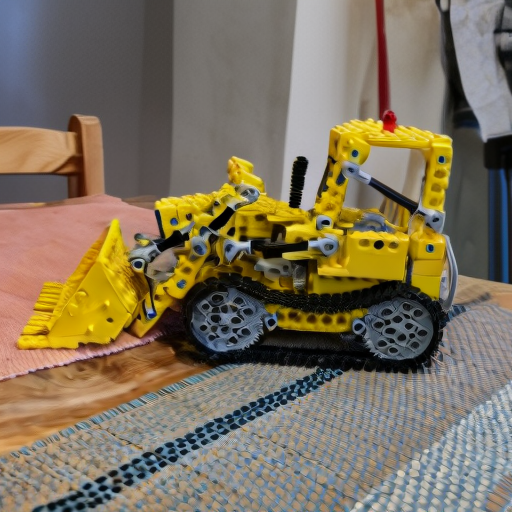}\\
        \includegraphics[width=\linewidth]{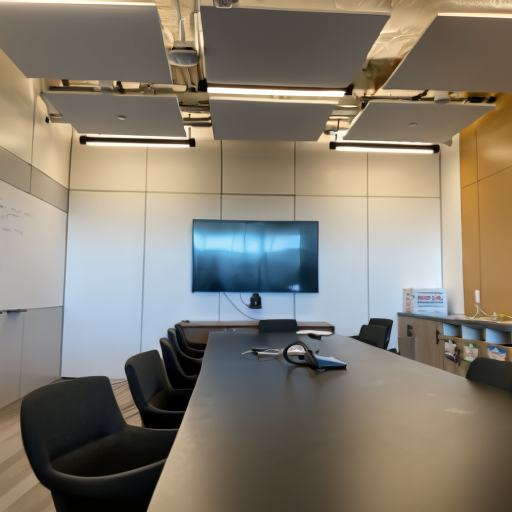}\\
        \includegraphics[width=\linewidth]{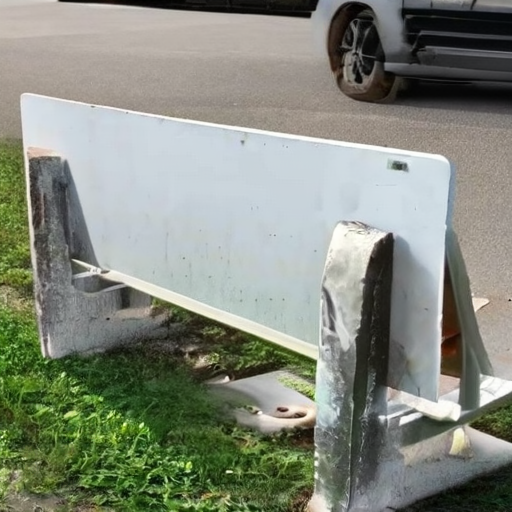}
        \caption{Diffusion model sample}
    \end{subfigure}
    \end{tabular}
    \caption{\textbf{PixelNeRF Visualization}. Here we show (a,c) a visualization of the $64\times 64$ RGB component of the PixelNeRF output, and (b,d) the corresponding sample from the diffusion model, which is conditioned on the PixelNeRF outputs.}
    \label{fig:pixelnerf}
\end{figure*}

\section{PixelNeRF Details}
\label{sec:suppixel}

Our PixelNeRF module is inspired by but not identical to the architecture proposed in the original work~\cite{yu2021pixelnerf}. The inputs are $N$ images along with their camera poses (extrinsic and intrinsic matrices), along with a target camera pose. The output is an approximate rendering at the target camera pose (both RGB and feature channels), which is concatenated with the input to the diffusion U-Net to provide a strong conditioning signal that encodes the pose and image content of the target novel view. During inference, we typically resize (but do not crop) the inputs to PixelNeRF such that their shorter dimension is $512$ pixels, since the model was trained on $512 \times 512$ resolution inputs. The output target image is always $64\times 64 \times 131$ ($3$ RGB channels plus $128$ feature channels), to match the latent resolution of the diffusion model.

The PixelNeRF module begins by passing all input images through a 2D U-Net to create feature images of equal spatial resolution with $128$ channels. We then cast rays through each pixel of the target image, and sample $128$ points along each ray from depth $0.5$ to $\infty$ (uniform up to distance $1$, then linear in disparity). We reproject these points into each of the input cameras and gather corresponding features from the feature images to make a gathered tensor of size $64 \times 64 \times N \times 128$. We append positionally-encoded 3D locations, as well as the mean and variance of these features over the $N$-long dimension corresponding to the number of inputs. 
A small MLP then processes this full tensor along the channel dimension to output a new set of features and weights. These weights are then used to compute a weighted sum along the $N$-long dimension, thereby producing a new tensor of size $64 \times 64 \times 128$. A second MLP then processes this summed tensor along the channel dimension to produce the final output of size $64 \times 64 \times (3 + 128)$.

All learned components (including the 2D U-Net used to extract image features) are initialized randomly and optimized jointly with the fine-tuned diffusion model U-Net. As mentioned in the main text, we apply an RGB reconstruction loss to encourage the PixelNeRF module to learn a useful conditioning signal. See Figure~\ref{fig:pixelnerf} for a visualization of our PixelNeRF model.

\section{Dataset Details}
\label{sec:suppdataset}

For LLFF and DTU, we use the standard train/test splits proposed by earlier works. For RealEstate10k and CO3D, we select the training views evenly from all the frames and use every 8th of the remaining frames for evaluation. For the mip-NeRF 360 dataset we design a heuristic to choose a train split of views that are uniformly distributed around the hemisphere and pointed toward the central object of interest: We randomly sample $10^6$ different 9-view splits and use the one that minimizes these heuristic losses, then further choose the 6- and 3-view splits to be subsets of the 9-view split.

We carefully rescale each dataset to be compatible with the near plane of $0.5$ expected by the PixelNeRF module. DTU, CO3D, mip-NeRF 360 are rescaled by setting the ``focus point'' of the data to the origin and rescaling camera positions to fit inside a  $[-1, 1]^3$ cube. RealEstate10k is pre-scaled by its creators to have a reasonable near distance of $1.0$, so we simply multiply its camera positions by $0.5$. LLFF similarly provides a near bound based on the COLMAP point cloud, which we use to rescale the data to allow a $0.5$ near plane. 

\section{Baselines Details}
\label{sec:suppbaseline}

Our Zip-NeRF~\cite{barron2023zip} baseline has slight hyperparameter modifications from the original that better suit few-view reconstruction. This was done primarily to provide a maximally competitive baseline for our model, but these same hyperparameters are used by our model as well, and we observe a modest performance improvement due to them.
In particular, we use: 
\begin{itemize}
    \item Distortion loss with weight $0.01$,
    \item Normalized weight decay on the NGP grid parameters with strength $0.1$,
    \item A smaller view-dependence network with width 32 and depth 1, to avoid overfitting,
    \item No hexagonal spiral control points, to accelerate rendering at the cost of introducing some aliasing,
    \item A downweighted density in the ``contracted'' region of space outside of the unit sphere, wherein we multiply the density emitted by Zip-NeRF by $\lft| \operatorname{det}\lft(\mathbf{J}_{\mathcal{C}}(\mathbf{x})\rgt) \rgt|$ (the isotropic scaling induced by the contraction function, see the supplement of \citet{barron2023zip}).
\end{itemize}
We find that this baseline performs competitively on forward-facing scenes such as LLFF and RealEstate10k, especially with 9 input views, but often produces many floaters or fails to reconstruct any meaningful geometry on the more difficult datasets. Further tuning and the addition of other heuristic regularizers (\eg, the techniques used in RegNeRF or FreeNeRF) would likely improve results. However, the point of this model is to show baseline reconstruction performance with our diffusion model regularizer disabled, rather than to be a state-of-the-art few-view method. To re-emphasize this: the \textbf{only} difference between results labeled ``Ours'' and ``Zip-NeRF'' is that the diffusion regularizer weight is set to 0, all other hyperparameters are identical.

For RegNeRF~\cite{niemeyer2022regnerf} and FreeNeRF~\cite{yang2023freenerf}, we use the result images shared by the authors for the LLFF and DTU datasets, and run the authors' code for FreeNeRF on the other three datasets.
For DiffusioNeRF~\cite{wynn2023diffusionerf}, we use the authors' result images on the DTU dataset, and run their code on the other four datasets.
Because SimpleNeRF~\cite{somraj2023simplenerf} used different train/test splits for LLFF and DTU, we run the authors' code on all five datasets.

SparseFusion~\cite{zhou2023sparsefusion} was originally trained on CO3D using masked images of foreground objects. We re-train their models using the unmasked images in a category-specific manner, then use their 3D distillation pipeline to obtain the final rendered images.

ZeroNVS~\cite{sargent2023zeronvs} is a concurrent work that trains a diffusion model for novel view synthesis of scenes from a single image. 
To evaluate its performance on multiview inputs, we modify its reconstruction pipeline by using the input view closest to the sampled random view for conditioning the diffusion model.
For DTU and mip-NeRF 360 scenes (which they evaluate in their paper for the single input case), we follow their viewpoint selection strategy for sampling new views.
For CO3D, we use the same strategy that is employed for the mip-NeRF 360 scenes.
For RealEstate10K, we sample new views on a spline path fitted from the input views, then perturb them.
For LLFF, we sample new views on a circle fitted from the input views, then perturb them.

\end{document}